\documentclass[10pt,twocolumn,letterpaper]{article}

\usepackage[numbers]{natbib}
\usepackage[resetlabels]{multibib}
\newcites{supp}{References}
\usepackage[pagebackref=true,breaklinks=true,colorlinks,bookmarks=false]{hyperref}
\usepackage{etoolbox}
\makeatletter
\patchcmd\Hy@backout{\@auxout}{\@mainaux}{}{\fail}
\patchcmd\Hy@backout{\@auxout}{\@mainaux}{}{\fail} 
\makeatother

\usepackage{iccv}
\usepackage{times}
\usepackage{epsfig}
\usepackage{graphicx}
\usepackage{amsmath}
\usepackage{amssymb}
\usepackage{booktabs}
\usepackage{enumitem}
\usepackage{dsfont}
\usepackage{algorithm, algorithmic}
\algsetup{linenosize=\small}
\usepackage{tablefootnote}
\usepackage[flushleft]{threeparttable} 
\usepackage{balance}
\usepackage{xcolor}
\usepackage{color,colortbl}
\usepackage{overpic}
\usepackage{wrapfig}
\usepackage[normalem]{ulem}

\usepackage[accsupp]{axessibility}

\definecolor{tabblue}{HTML}{00b4d8}

\input{insbox.tex}
\newcommand{\name}{HiFace}
\newcommand{\module}{SD-DeTail}

\usepackage{wasysym}

\usepackage{multirow}
\usepackage{multicol}
\usepackage{soul}
\setul{0.5ex}{0.3ex}

\newcommand{\myparagraph}[1]{\vspace{2mm}\noindent\textbf{#1}}

\definecolor{col_nose}{RGB}{181,228,140}
\definecolor{col_mouth}{RGB}{255,183,3}
\definecolor{col_forehead}{RGB}{189,178,255}
\definecolor{col_cheek}{RGB}{144,224,239}
\definecolor{col_table}{RGB}{175,227,246} 

\definecolor{mycolor1}{rgb}{0.85000,0.32500,0.09800}%
\definecolor{mycolor2}{rgb}{0.92900,0.69400,0.12500}%
\definecolor{mycolor3}{rgb}{0.49400,0.18400,0.55600}%
\definecolor{mycolor4}{rgb}{0.87843,0.76471,0.98824}%
\definecolor{mycolor5}{rgb}{0.46600,0.67400,0.18800}%
\definecolor{mycolor6}{rgb}{0.30100,0.74500,0.93300}%
\definecolor{mycolor7}{rgb}{0.00000,0.44700,0.74100}%

\definecolor{best_two}{RGB}{72,149,239} 
\definecolor{best}{RGB}{179,11,0} 

\usepackage[capitalize]{cleveref}
\crefname{section}{Sec.}{Secs.}
\Crefname{section}{Section}{Sections}
\Crefname{table}{Table}{Tables}
\crefname{table}{Tab.}{Tabs.}

\iccvfinalcopy 

\ificcvfinal\fi

\begin{document}

\title{{\name}: High-Fidelity 3D Face Reconstruction by\\Learning Static and Dynamic Details}

\author{Zenghao Chai$^{1, 2}$\thanks{Work done when the author was an intern at MSRA.} ~~~~ Tianke Zhang$^{2}$ ~~~~ Tianyu He$^{3}$ ~~~~ Xu Tan$^{3}$\thanks{Corresponding author: Xu Tan (xuta@microsoft.com).} ~~~~ Tadas Baltru\v{s}aitis$^{4}$\\ HsiangTao Wu$^{5}$ ~~~~ Runnan Li$^{5}$ ~~~~ Sheng Zhao$^{5}$ ~~~~ Chun Yuan$^{2}$ ~~~~ Jiang Bian$^{3}$ \\
$^{1}$National University of Singapore~~~~ $^{2}$Tsinghua University  ~~~~~
$^{3}$Microsoft Research Asia \\$^{4}$Microsoft Mixed Reality \& AI Lab ~~~~~ $^{5}$Microsoft Cloud + AI \\
\tt\small zenghaochai@gmail.com ~~ ztk21@mails.tsinghua.edu.cn ~~ yuanc@sz.tsinghua.edu.cn \\ 
\tt\small \{tianyuhe,xuta,tabaltru,musclewu,runnan.li,sheng.zhao,jiang.bian\}@microsoft.com
}

\maketitle
\ificcvfinal\thispagestyle{empty}\fi

\begin{abstract}
    3D Morphable Models (3DMMs) demonstrate great potential for reconstructing faithful and animatable 3D facial surfaces from a single image.
    The facial surface is influenced by the coarse shape, as well as the static detail ({\eg}, person-specific appearance) and dynamic detail ({\eg}, expression-driven wrinkles). Previous work struggles to decouple the static and dynamic details through image-level supervision, leading to reconstructions that are not realistic. 
    In this paper, we aim at high-fidelity 3D face reconstruction and propose {\name} to explicitly model the static and dynamic details. 
    Specifically, the static detail is modeled as the linear combination of a displacement basis, while the dynamic detail is modeled as the linear interpolation of two displacement maps with polarized expressions. 
    We exploit several loss functions to jointly learn the coarse shape and fine details with both synthetic and real-world datasets, which enable {\name} to reconstruct high-fidelity 3D shapes with animatable details.
    Extensive quantitative and qualitative experiments demonstrate that {\name} presents state-of-the-art reconstruction quality and faithfully recovers both the static and dynamic details. 
    Our project page: \href{https://project-hiface.github.io}{https://project-hiface.github.io}.   
\end{abstract}

\begin{figure}[t!]
\begin{overpic}[trim=0cm 0cm 0cm 0cm,clip,width=1\linewidth,grid=false]{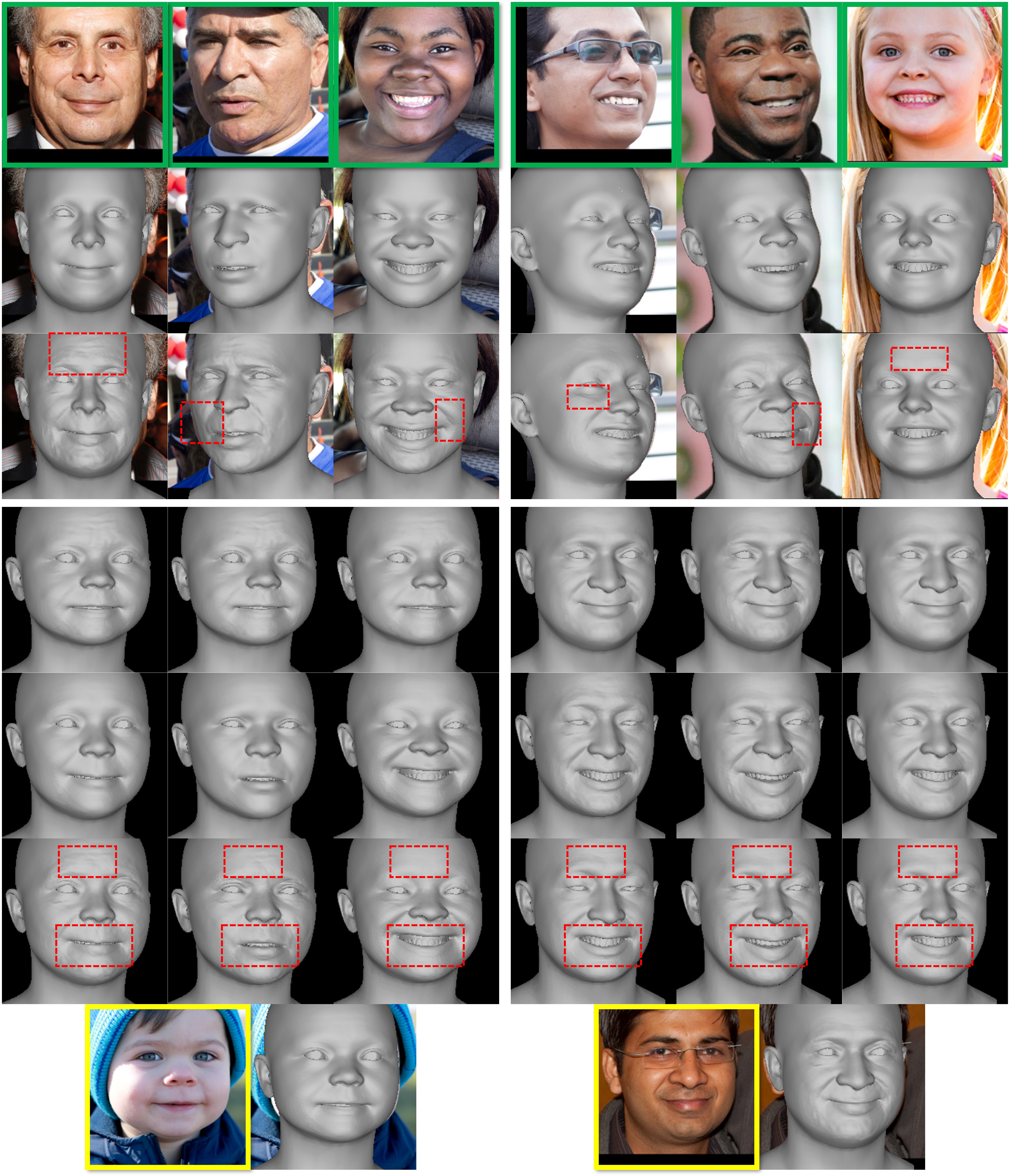}
\end{overpic}
\vspace{-5pt}
\caption{We propose {\name} to reconstruct high-fidelity 3D face with realistic and animatable details. \textbf{Reconstruction}: given a single image (1st-row), {\name} faithfully reconstructs a coarse shape (2nd-row) with vivid details (3rd-row). \textbf{Animation}: given a source face (yellow box), {\name} can animate the static (4th-row), dynamic (5th-row), or both (6th-row) details of the driving images (green box). Images are taken from FFHQ~\cite{karras2019style} and CelebA~\cite{CelebAMask-HQ}.
}
\vspace{-10pt}
\label{Fig.teaser}
\end{figure}

\section{Introduction}
The reconstruction of a 3D face from a single image has drawn much attention recently~\cite{tewari2017mofa,deng2019accurate,lee2020uncertainty,MICA}. It has tremendous potential applications like face recognition~\cite{cao2018pose,schroff2015facenet,blanz2002face,romdhani2002face}, face animation~\cite{cudeiro2019capture,yue2022anifacegan}, virtual reality~\cite{bouaziz2013online,ribera2017facial,hu2017avatar}, {\etc} For example, the reconstructed 3D face representation can be driven by an audio~\cite{cudeiro2019capture}, or a video from another person~\cite{kim2018deep}.

To build a flexible and animatable facial representation, a popular way is to leverage the success of 3D Morphable Models (3DMMs)~\cite{blanz1999morphable,booth2018large,cao2013facewarehouse,li2017learning,smith2020morphable}, which decouple the influence of shape, expression, albedo, and others by modeling them in separate coefficients.
Typically in literature, one can achieve coarse shape reconstruction in coefficients-fitting optimization~\cite{gecer2019ganfit,wood2022dense,bao2021high,yang2020facescape,bai2022ffhq}, or an analysis-by-synthesis pipeline~\cite{tewari2017mofa,deng2019accurate,MICA,luo2021normalized}. As 3DMMs typically capture only the coarse facial geometry and are not capable of representing fine details (\eg, wrinkles), recent advances model such details with a displacement map~\cite{chen2020self,yang2020facescape,chen2019photo,cao2015real,yamaguchi2018high}.
However, previous work fails to model the distinction between static and dynamic factors of fine detail, leading to errors in reconstructions.
For example, given that one may drive the expression of a young man from an old man, trivially transferring all wrinkles from the old man to the young man could make the young man look unnatural.
In this sense, Feng {\etal}~\cite{feng2021learning} implicitly leverages the person-specific identity and expression as conditions to generate the details. Although effective, they optimize the model in an analysis-by-synthesis pipeline with only the image-level supervision, leading to insufficient decoupling of static and dynamic details and inconsistent animation results (see Fig.~\ref{fig:animation}).

Therefore, we propose {\name} to explicitly model the static and dynamic details for high-fidelity 3D face reconstruction, by designing {\module} module to decouple the static and dynamic factors.
More specifically, for person-specific static detail, instead of directly predicting the displacement map that may increase the difficulty of detail prediction~\cite{feng2021learning,danvevcek2022emoca}, we follow the spirit of 3DMMs to build a displacement basis from the captured facial scans with age diversity~\cite{raman2022mesh,wood2021fake}. In this way, the model is trained to predict the coefficients of the displacement basis, and make the detail prediction easier. For dynamic detail, since it is highly expression-dependent, directly modeling it with one displacement basis is quite difficult. Therefore, based on the fact that the expression can be interpolated by a compressed and a stretched expressions~\cite{raman2022mesh}, we build two displacement bases for the compressed and stretched expressions from the captured scans respectively, and learn to regress the displacement coefficients with the ground-truth labels. Therefore, we can obtain the dynamic detail by linearly interpolating the compressed and stretched displacement maps, which are derived from the displacement bases and the predicted coefficients. Finally, the predicted static and dynamic details are merged with the coarse shape to formulate the final output.

Since we would like the final output to contain both the coarse shape and high-frequency detail, we propose several novel loss functions to learn coarse shape and details simultaneously from both the synthetic and real-world datasets. For details, we leverage the ground-truth static and dynamic displacement maps of the synthetic dataset~\cite{wood2021fake,raman2022mesh} as supervision. While for the coarse shape, we leverage the ground-truth vertex of the synthetic dataset as supervision. We also follow the previous methods~\cite{feng2021learning,deng2019accurate,wood2022dense} to leverage self-supervised losses for all training images.

Overall, with the above insights and techniques, {\name} enables the reconstruction of high-fidelity 3D faces from a single image, and decouples static and dynamic details that are naturally animatable (see Fig.~\ref{Fig.teaser}). 
We demonstrate that the proposed {\name} reconstructs realistic and faithful 3D faces, reaching state-of-the-art performance both quantitatively and qualitatively. In addition, {\name} is compatible with optimization-based methods~\cite{wood2022dense}, and is flexible to transfer vivid expressions and details from one person to another.
In summary, our contributions are:
\begin{itemize}[leftmargin=*,nosep,nolistsep]
\item We propose {\name} to model the static and dynamic details explicitly, and demonstrate the benefits of synthetic data in decoupling the static and dynamic factors for detailed 3D face reconstruction.
\item We propose novel loss functions in {\name} to learn 3D representations of coarse shape and fine details simultaneously from both the synthetic and real-world images.
\item We achieve state-of-the-art reconstruction quality both quantitatively and qualitatively, with over $15\%$ performance gains in the region-aware benchmark~\cite{REALY}.
\item We show that our {\module} is easy to plug-and-play into optimization-based methods and can transfer expressions and details from one to another for face animation.
\end{itemize}

\section{Related Work}

3D face reconstruction from monocular images has received much attention in the past decades.
Among them, 3D Morphable Models (3DMMs) are widely used to build 3D representations.
Below we review the works that are related to them, and a full in-depth review can be found in recent surveys~\cite{zollhofer2018state,morales2021survey,egger20203d}.

\myparagraph{3D Morphable Model} (3DMMs)~\cite{egger20203d} are statistical models widely used to constrain the distribution of 3D faces. The seminal work~\cite{blanz1999morphable} presents $200$ scans to generate shape and texture bases with Principal Component Analysis (PCA)~\cite{abdi2010principal}, and formulate 3DMMs as linear models by the generated bases.
After that, expression models~\cite{vlasic2006face,li2017learning,cao2013facewarehouse} are proposed to support face manipulation. Recent advances~\cite{REALY,ploumpis2019combining,smith2020morphable,dai20173d,li2020learning} are proposed to expand the expressiveness of 3DMMs and play a crucial role in 3D face reconstruction. 3DMMs make it possible to simplify the 2D-to-3D problem into a regression task, which typically presents an analysis-by-synthesis fashion to estimate the coefficients of 3DMMs. 
In this paper, we follow the spirit of the 3DMMs family to present the decoupled static and dynamic details for 3D face reconstruction.

\begin{figure*}[ht!]
    \centering
    \begin{overpic}[trim=0cm 0cm 0cm 0cm,clip,width=1\linewidth,grid=false]{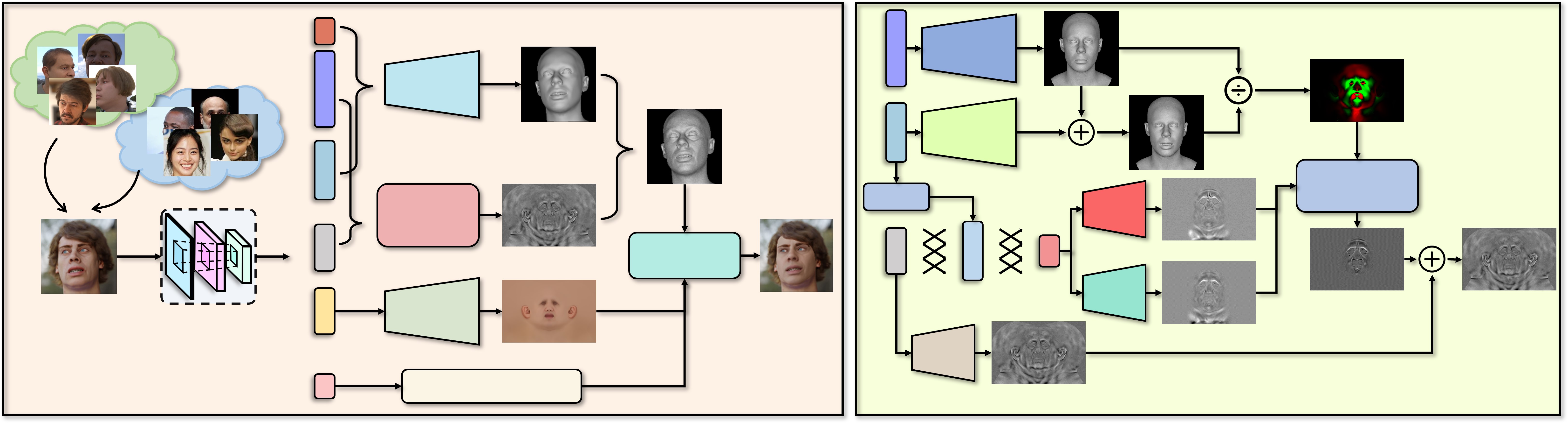}
    \end{overpic}
    \put(-315,123){\footnotesize (a). Overview of {\name}}
    \put(-100,123){\footnotesize (b). Overview of {\module}}
    \put(-481,34){\bfseries\scriptsize Input}
    \put(-481,28){\bfseries\scriptsize Image}
    \put(-445,30){\scriptsize ResNet-50}
    \put(-490,80){\bfseries\scriptsize Synthetic}  
    \put(-470,70){\bfseries\scriptsize Real World} 
    \put(-405,10){\bfseries\scriptsize $\boldsymbol{\gamma}$}
    \put(-405,33){\bfseries\scriptsize $\boldsymbol{\alpha}$}
    \put(-405,53){\bfseries\scriptsize $\boldsymbol{\varphi}$}
    \put(-405,78){\bfseries\scriptsize $\boldsymbol{\xi}$}
    \put(-405,103){\bfseries\scriptsize $\boldsymbol{\beta}$}
    \put(-405,122){\bfseries\scriptsize $\mathbf{p}$}
    \put(-374,105){\scriptsize 3D Model}
    \put(-376,64){\scriptsize {\module}}    
    \put(-365,34){\scriptsize $\mathbf{B}_{\text{alb}}$}    
    \put(-368,9){\scriptsize Illumination Model}       
    \put(-340,90){\scriptsize Coarse Shape}       
    \put(-352,49.5){\scriptsize Displacement Map} 
    \put(-340,19.5){\scriptsize Albedo Map} 
    \put(-293,51){\scriptsize Renderer} 
    \put(-298,69){\scriptsize Detail Shape}     
    \put(-260,34){\bfseries\scriptsize Rendered}
    \put(-255,28){\bfseries\scriptsize Image}    
    \put(-222.5,115){\bfseries\scriptsize $\boldsymbol{\beta}$}
    \put(-222.5,88){\bfseries\scriptsize $\boldsymbol{\xi}$}
    \put(-222.5,69){\bfseries\scriptsize \textit{AdaIN}}
    \put(-222.5,51){\bfseries\scriptsize $\boldsymbol{\varphi}$} 
    \put(-195,116){\scriptsize $\mathbf{B}_{\text{id}}$}
    \put(-195,89){\scriptsize $\mathbf{B}_{\text{exp}}$}
    \put(-206,37){\scriptsize MLP}
    \put(-183,37){\scriptsize MLP}
    \put(-168,42){\bfseries\scriptsize $\boldsymbol{\phi}$}
    \put(-203,20){\scriptsize $\mathbf{B}_{\text{sta}}$}
    \put(-184,6){\scriptsize Static Detail}
    \put(-142,110){\scriptsize $\mathbf{S}_{\text{neu}}$}
    \put(-115,80){\bfseries\scriptsize $\mathbf{S}$}
    \put(-102,95){\scriptsize Eq.~\ref{Eq.tension}}
    \put(-44,73){\scriptsize Eq.~\ref{Eq.dyn_detail}}
    \put(-151,65){\scriptsize $\mathbf{B}_{\text{com}}$}
    \put(-149,39){\scriptsize $\mathbf{B}_{\text{str}}$}
    \put(-130,52){\scriptsize Compressed}
    \put(-127,25){\scriptsize Stretched}
    \put(-85,90){\scriptsize Vertex Tension $\mathbf{M}_{\text{uv}}$}
    \put(-82,73){\bfseries\scriptsize \textit{Interpolate}}
    \put(-90,36){\scriptsize Dynamic Detail}
    \put(-29,36){\bfseries\scriptsize Output}
    \put(-420,-10){\bfseries\scriptsize$\mathbf{p}$: pose coef \quad $\boldsymbol{\beta}$: identity coef \quad $\boldsymbol{\xi}$: expression coef \quad $\boldsymbol{\varphi}$: static coef \quad $\boldsymbol{\alpha}$: albedo coef \quad $\boldsymbol{\gamma}$: SH coef \quad $\boldsymbol{\phi}$: dynamic coef}
    \vspace{5pt}
    \caption{\textbf{Illustration of {\name}}. (a). Learning framework of {\name}. Given a monocular image, we regress its shape and detail coefficients to synthesize a realistic 3D face, and leverage a differentiable renderer~\cite{genova2018unsupervised} to train the whole model end-to-end from synthetic~\cite{wood2021fake,raman2022mesh} and real-world~\cite{CelebAMask-HQ,mollahosseini2017affectnet} images. (b). The pipeline of \textbf{S}tatic and \textbf{D}ynamic \textbf{De}coupling for De\textbf{Tail} Reconstruction ({\module}). We explicitly decouple the static and dynamic factors to synthesize realistic and animatable details. Given the shape and static coefficients, we regress the static and dynamic details through displacement bases and interpolate them into the final details through vertex tension~\cite{raman2022mesh}.}
    \label{fig:pipeline}
    \vspace{-10pt}
\end{figure*}
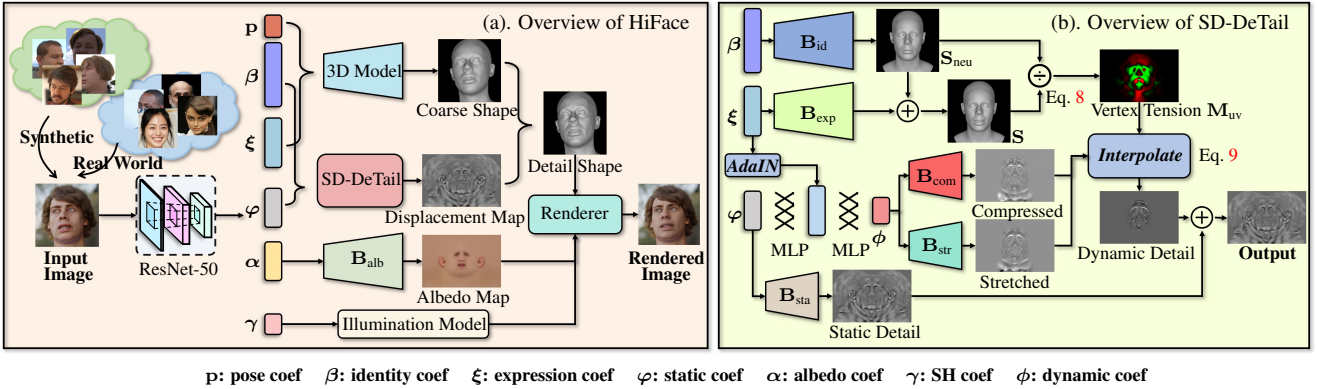

\myparagraph{Coarse Shape Reconstruction.} 
Traditional optimization-based methods~\cite{gecer2019ganfit,wood2022dense,bao2021high,yang2020facescape,bai2022ffhq} directly optimize the 3DMM coefficients of given 2D images.
While such methods work well in controlled settings ({\eg}, frontal view, no occlusion), they heavily rely on high-quality annotations. 
Learning-based methods leverage the advances of CNNs~\cite{tewari2017mofa,deng2019accurate,danvevcek2022emoca,sanyal2019learning} and GCNs~\cite{lin2020towards,lee2020uncertainty,gao2020semi} to learn high-level representations from large-scale images in the wild. These methods show plausible generalization over diverse environments.
To train the network end-to-end, recent methods leverage the differentiable renderers~\cite{genova2018unsupervised,dib2021towards,zhu2020reda,liu2019soft}, along with the photo loss, perceptual loss, and landmark loss~\cite{deng2019accurate,danvevcek2022emoca,genova2018unsupervised,tewari2017mofa,wen2021self} to optimize the network in a self-supervised manner. Different from these coarse shape reconstruction methods, we aim at high-fidelity 3D face reconstruction with both coarse shape and fine details.

\myparagraph{Detail Reconstruction.}
While 3DMMs can reconstruct coarse 3D face shapes from 2D images, they struggle with reconstructing fine-level details, such as forehead wrinkles and crows-feet. To fill this gap, shape by shading (SfS)~\cite{jiang20183d,li2018feature,garrido2016reconstruction,suwajanakorn2014total} methods reconstruct the facial details from monocular images or videos. However, these methods are sensitive to occlusions and large poses. Recent advances~\cite{feng2021learning,ling2022structure,chen2020self,chen2019photo,lei2023Ahierarchical} leverage displacement maps to present details. These methods explicitly re-topologize the coarse shape and present residual bias to generate geometric details.
The main challenge of detail reconstruction is the difficulty in learning the nuances and disentangling the static and dynamic details from only self-supervised learning. Ground-truth labels of the details are helpful to guide the learning process. However, it is difficult to obtain such fine-grained labels on real data.

\myparagraph{Synthetic Dataset.}
Several methods~\cite{MICA,yang2020facescape,jackson2017large,tuan2017regressing,martyniuk2022dad} utilize rendered faces or fitted coefficients to synthesize 3D-2D pairs. These ground-truth pairs lack diversity over background, illumination, and assets, making them hard to generalize well to real-world images.
Recent advances in synthetic data generation~\cite{wood2021fake,raman2022mesh} demonstrate its ability to generalize to real-world settings, and diverse to compensate for the domain gap to real-world images. In this paper, we leverage high-quality data with ground-truth labels to explore the detailed 3D face reconstruction.

\section{Methodology}

\subsection{Preliminary}

We adopt a common practice~\cite{deng2019accurate,feng2021learning} to represent a textured coarse shape with a 3D face model, an illumination model, and a camera model.

\myparagraph{3D Face Model.}
The 3D shape $\mathbf{S}$ and albedo $\mathbf{A}$ are represented by:
\begin{equation}
\begin{aligned}
\mathbf{S} &= \bar{\mathbf{S}}+\boldsymbol{\beta} \mathbf{B}_{\text{id}}+\boldsymbol{\xi} \mathbf{B}_{\text{exp}} \\
\mathbf{A} &= \bar{\mathbf{A}}+\boldsymbol{\alpha} \mathbf{B}_{\text{alb}}
\end{aligned},
\label{Eq.shape3dmm}
\end{equation}
where $\bar{\mathbf{S}}$ and $\bar{\mathbf{A}}$ are the mean shape and albedo. $\mathbf{B}_{\text{id}}$, $\mathbf{B}_{\text{exp}}$, and $\mathbf{B}_{\text{alb}}$ are bases~\cite{wood2022dense} of $256$-dim identity, $233$-dim expression, and $300$-dim albedo, respectively.
The coarse shape $\mathbf{S}$ in the bind pose is deformed from a neutral shape $\mathbf{S}_{\text{neu}}=\bar{\mathbf{S}}+\boldsymbol{\alpha} \mathbf{B}_{\text{id}}$ with expression component $\boldsymbol{\xi}\mathbf{B}_{\text{exp}}$.
$\boldsymbol{\beta}$, $\boldsymbol{\xi}$, and $\boldsymbol{\alpha}$ are the corresponding identity, expression, and albedo coefficients for generating a coarse shape. In this paper, the coarse shape $\mathbf{S}$ contains $n_v=7,667$ vertices and $n_f=14,832$ triangles with $512\times512\times3$ albedo.

\myparagraph{Pose \& Camera Model.}
To estimate the face pose, we follow~\cite{wood2021fake,wood2022dense} to predict skeletal pose $\mathbf{p}=[\boldsymbol{\theta}| \mathbf{t}]$, where $\boldsymbol{\theta} \in \mathds{R}^{3j}$ and $\mathbf{t} \in \mathds{R}^{3}$ are the local joint rotations and root joint translation, respectively. $j=4$ indicates $4$ skeletal joints {\wrt} the head, neck, and two eyes.
We perform a standard linear blend skinning (LBS) function~\cite{lewis2000pose} (with per-vertex weights $\mathbf{W}\in \mathds{R}^{j\times n_v}$) to rotate $\mathbf{S}$ about joint locations $\mathbf{J}\in\mathds{R}^{3j}$ by $\mathbf{p}$ to obtain $\mathbf{S}_{\mathbf{p}}$:
\begin{equation}
    \mathbf{S}_{\mathbf{p}}=\text{LBS}(\mathbf{S},\mathbf{p}, \mathbf{J};\mathbf{W}),
\end{equation}
where $\mathbf{J}$ is the joint locations in the bind pose determined by $\mathbf{J}=\mathcal{J}(\boldsymbol{\beta}): \mathds{R}^{|\boldsymbol{\beta}|}\rightarrow \mathds{R}^{3j}$.
Then we use an orthographic camera model to project 3D vertices in $\mathbf{S}_{\mathbf{p}}$ to the 2D plane.

\myparagraph{Illumination Model.}
We follow previous work~\cite{deng2019accurate} to use Spherical Harmonics (SH)~\cite{ramamoorthi2001efficient} to estimate the illumination of a given image. The shaded texture $\mathbf{T}$ is computed as:
\begin{equation}
    \mathbf{T} =  \mathbf{A}\odot \sum\nolimits_{k=1}^{9}\boldsymbol{\gamma}_k \mathbf{\boldsymbol{\Psi}}_{k} (\mathbf{N}),
    \label{Eq.texture}
\end{equation}
where $\odot$ denotes the Hadamard product, $\mathbf{N}$ is the surface normal of $\mathbf{S}$ in UV coordinates, $\boldsymbol{\Psi}: \mathds{R}^{3}\rightarrow \mathds{R}$ are SH basis function and $\boldsymbol{\gamma}\in \mathds{R}^9$ is the corresponding SH coefficient.

\subsection{Overview of {\name}}
\vspace{-2pt}

\myparagraph{Key Idea.}
The key idea of {\name} is to explicitly model the static (\eg, person-specific properties) and dynamic ({\eg}, expression-driven wrinkles) details, allowing the model to reconstruct a high-fidelity 3D face from a single image with realistic and animatable details.

\myparagraph{Overview.}
The goal of {\name} is to reconstruct 3D shapes with realistic details from a single image. The overview of {\name} is illustrated in Fig.~\ref{fig:pipeline}(a).
We leverage a feature extractor (\ie, ResNet-50~\cite{he2016deep}) to regress corresponding coefficients from an input image.
Our model jointly predicts both the coarse-level shapes and the fine-level details. For coarse-level shapes, we regress shape parameters ({\ie}, identity, expression, albedo, illumination, and pose) of a parametric face model. For the fine-level details, we propose a novel way to model it through the separation of static and dynamic factors and formulate the generation of details into the problems of 3DMM coefficients regression and displacement maps interpolation.

Note that the facial details are based on the coarse shape, we thereby exploit novel loss functions to learn 3D representations of coarse shape and details simultaneously from the synthetic dataset with ground-truth labels. To generalize our model to real-world images, we also present several self-supervised losses to train the model with both synthetic data and real-world images coherently.
As a result, {\name} can faithfully reconstruct the facial details of a given image, or animate a face by combining the decoupled static and dynamic coefficients that come from different individuals.

\subsection{Decoupling Static and Dynamic Details}

We propose \textbf{S}tatic and \textbf{D}ynamic \textbf{De}coupling for De\textbf{Tail} Reconstruction ({\module}).
The facial details are basically composed of a static factor and a dynamic factor:
\begin{equation}
    \mathbf{D} = \mathbf{D}_{\text{sta}} + \mathbf{D}_{\text{dyn}},
    \label{Eq.outdisp}
\end{equation}
where $\mathbf{D}_{\text{sta}}$ and $\mathbf{D}_{\text{dyn}}$ indicate details from static and dynamic factors, respectively.

Concretely, the static factor is the inherent property of the identity ({\ie}, the given 2D face), and originates from the appearance and age attributes. As for the dynamic factor, it is typically driven by the expression and influenced by person-specific properties.

\myparagraph{Static Detail Generation.}
To simplify the problem, we are inspired by 3DMMs, which parameterize the statistical models to simplify the 2D-to-3D problem. We build a $300$-dim displacement basis $\mathbf{B}_{\text{sta}}$ from the captured $332$ scans~\cite{raman2022mesh} by PCA~\cite{abdi2010principal}. The scans contain diverse age groups in a neutral expression. Then we regress the coefficient $\boldsymbol{\varphi}$ to synthesize the static detail $\mathbf{D}_{\text{sta}}$ from the image:
\begin{equation}
\begin{aligned}
\mathbf{D}_{\text{sta}} =  \bar{\mathbf{D}}_{\text{sta}}+\mathbf{\boldsymbol{\varphi}} \mathbf{B}_{\text{sta}}
\end{aligned},
\label{Eq.age3dmm}
\end{equation}
where $\bar{\mathbf{D}}_{\text{sta}}$ and $\mathbf{B}_{\text{sta}}$ are the mean displacement map and displacement basis for static details, respectively.

\myparagraph{Dynamic Detail Generation.}
Due to the high diversity and complexity of expression representation, directly generating dynamic details from expression is quite difficult.
Therefore we simplify the expression representation by using an interpolation between two displacement maps: compressed and stretched~\cite{raman2022mesh}.
For example, the compressed expression may indicate a state of frowning to the extreme, while the stretched expression may indicate a state of complete relaxation between the eyebrows. Other states of this area can be interpolated by these two polarized states.

Consequently, we generate the dynamic details through compressed and stretched displacement maps. Again, we build $26$-dim compressed $\mathbf{B}_{\text{com}}$ and stretched $\mathbf{B}_{\text{str}}$ displacement bases by PCA~\cite{abdi2010principal} to simplify the generation of displacement maps.
To generate the dynamic coefficients $\boldsymbol{\phi}=\{\boldsymbol{\phi}_{\text{com}},\boldsymbol{\phi}_{\text{str}}\}$, we apply the expression coefficient $\boldsymbol{\xi}$ into the static coefficient $\boldsymbol{\varphi}$ through AdaIN~\cite{huang2017arbitrary}, followed by the MLP transformation $\boldsymbol{\Phi}$ to obtain $\boldsymbol{\phi}$:
\begin{equation}
    \boldsymbol{\phi} = \boldsymbol{\Phi}\bigg(\sigma\big(\tilde{\boldsymbol{\xi}}\big)\Big(\frac{\boldsymbol{\varphi}-\mu(\boldsymbol{\varphi})}{\sigma(\boldsymbol{\varphi})}+\mu\big(\tilde{\boldsymbol{\xi}}\big)\Big)\bigg)
    \label{Eq.adain},
\end{equation}
where $\tilde{\boldsymbol{\xi}}$ is the affined vector from $\boldsymbol{\xi}$ via MLP transformation. $\mu$ and $\sigma$ indicate the mean and standard deviation.
$\boldsymbol{\phi}_{\text{com}}$ and $\boldsymbol{\phi}_{\text{str}}$ are coefficients for compressed and stretched displacement maps respectively.

Similar to Eq.~\ref{Eq.age3dmm}, the compressed and stretched displacement maps are formulated as:
\begin{equation}
\begin{aligned}
\mathbf{D}_{\text{com}} &=  \bar{\mathbf{D}}_{\text{com}}+\boldsymbol{\phi}_{\text{com}} \mathbf{B}_{\text{com}} \\
\mathbf{D}_{\text{str}} &=  \bar{\mathbf{D}}_{\text{str}}+\boldsymbol{\phi}_{\text{str}} \mathbf{B}_{\text{str}}
\end{aligned},
\label{Eq.exp3dmm}
\end{equation}
where $\bar{\mathbf{D}}_{\text{com}}$ and $\mathbf{B}_{\text{com}}$ are the mean displacement map and $26$-dim displacement basis for compressed detail, and $\bar{\mathbf{D}}_{\text{str}}$ and $\mathbf{B}_{\text{str}}$ are the mean displacement map and $26$-dim displacement basis for stretched detail, respectively.

Considering the coarse shape $\mathbf{S}$ can be obtained by deforming the neutral shape $\mathbf{S}_{\text{neu}}$ with the expression component $\boldsymbol{\xi}\mathbf{B}_{\text{exp}}$, such expression-driven deformation over face shape yields the ``tension'' over each vertex~\cite{raman2022mesh}, which influences facial details from expression.
Since $\mathbf{S}_{\text{neu}}$ and $\mathbf{S}$ posses the same topology, for each vertex $\mathbf{v}_i\in \mathbf{S}$ with $K$-edges $E_{i}=\{e_1,\cdots,e_{K}\}$ connected with $\mathbf{v}_i$, $E'_{i}=\{e'_1,\cdots,e'_{K}\}$ are the corresponding edges in $\mathbf{S}_{\text{neu}}$ that are connected to $\mathbf{v}'_i$. Then the tension at $\mathbf{v}_i$ is:
\begin{equation}
t_{\mathbf{v}_i}=1-\frac{1}{K}\sum\nolimits_{k=1}^{K} \frac{ \| e_k  \| }{\| e'_k  \| },
\label{Eq.tension}
\end{equation}
where $\| \cdot \|$ represents the edge length. Positive values of $t_{\mathbf{v}_i}$ indicate compression, negative values indicate stretch, and $0$-value indicates no change, respectively.

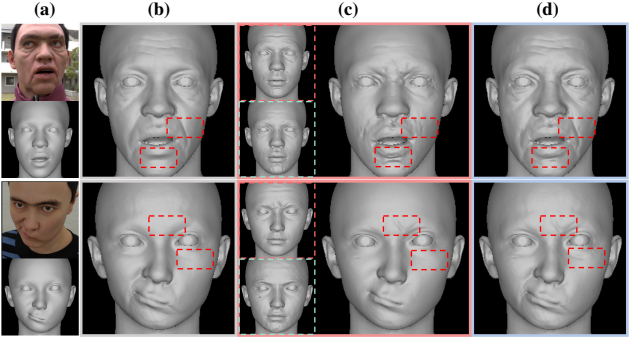
\begin{figure}
    \centering
    \begin{overpic}[trim=0cm 0cm 0cm 0cm,clip,width=1\linewidth,grid=false]{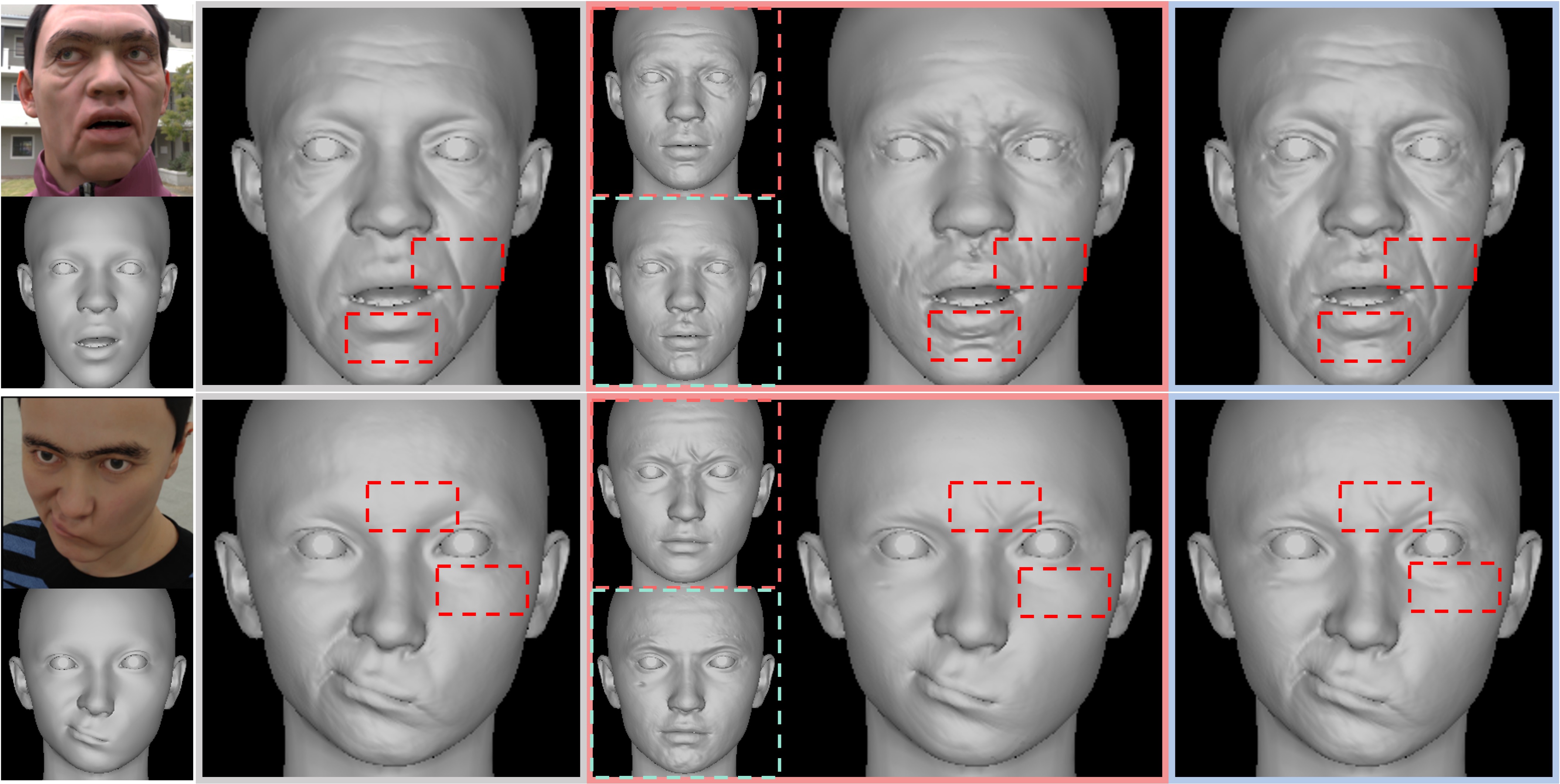}
    \end{overpic}
    \put(-225,122){\bfseries\scriptsize (a)}
    \put(-182,122){\bfseries\scriptsize (b)}
    \put(-110,122){\bfseries\scriptsize (c)}
    \put(-35,122){\bfseries\scriptsize (d)}
    \vspace{-5pt}
    \caption{\textbf{Illustration of displacement map composition in {\module}.} Given (a). an image (top) to reconstruct its coarse shape (bottom), we formulate the detail as (b). a static factor and (c). a dynamic factor interpolated by polarized states {\wrt} compressed (top) and stretched (bottom). (d). the output displacement map is linearly combined by (b) and (c) to present vivid details.}
    \label{fig:detail_example}
    \vspace{-10pt}
\end{figure}

The vertex tension $t_{\mathbf{v}_i}$ in $\mathbf{S}$ composes the tension map $\mathbf{M}_{\text{uv}}$ in UV coordinates. Then, the displacement map of the dynamic detail is the linear interpolation of $\mathbf{D}_{\text{com}}$ and $\mathbf{D}_{\text{str}}$:
\begin{equation}
    \mathbf{D}_{\text{dyn}} = \mathbf{M}_{\text{uv}}^{+} \odot \mathbf{D}_{\text{com}} + \mathbf{M}_{\text{uv}}^{-}\odot \mathbf{D}_{\text{str}},
    \label{Eq.dyn_detail}
\end{equation}
where $\mathbf{M}_{\text{uv}}^{+}$ and $\mathbf{M}_{\text{uv}}^{-}$ indicate the positive and negative value of $\mathbf{M}_{\text{uv}}$, respectively. Fig.~\ref{fig:detail_example} shows the effectiveness of {\module}. The dynamic factor interpolated by two polarized states introduces expression-related details and further decorates the static detail, yielding the final vivid output.

\subsection{Overall Loss Functions}

We propose several loss functions to train {\name} end-to-end. Specifically, we use static and dynamic detail losses to supervise the synthesized displacement maps from $\boldsymbol{\varphi}$ and $\boldsymbol{\phi}$. In addition, we leverage the coarse shape loss to supervise the reconstructed shape from $\boldsymbol{\beta}$ and $\boldsymbol{\xi}$. Finally, we follow previous methods~\cite{deng2019accurate,feng2021learning,wood2022dense} to leverage the differentiable renderer~\cite{genova2018unsupervised} to map the generated 3D shape into 2D images, by combining $\boldsymbol{\alpha}$, $\boldsymbol{\beta}$, $\boldsymbol{\xi}$, $\boldsymbol{\gamma}$, $\mathbf{p}$, $\boldsymbol{\varphi}$, $\boldsymbol{\phi}$. Then, we perform self-supervised losses to train in both synthetic and real-world images. See more details in the supplementary.

\myparagraph{Static and Dynamic Detail Losses.}
To explicitly train the details of each component, we leverage the ground-truth annotations from the synthetic dataset~\cite{raman2022mesh,wood2021fake} as supervision to assist the training process of our model. Specifically, we calculate the detail losses by estimating the $l_2$ distance between the reconstructed displacement maps and ground-truth {\wrt} static, compressed, and stretched components, and summarize them as $\mathcal{L}_{\text{detail}}$:
\begin{equation}
\begin{aligned}
     &\mathcal{L}_{\text{sta}}=\big\| \mathbf{M}_{\text{detail}}\odot (\mathbf{D}_{\text{sta}}-\hat{\mathbf{D}}_{\text{sta}}) \big\|_2 \\
     &\mathcal{L}_{\text{com}}=\big\| \mathbf{M}_{\text{detail}}\odot (\mathbf{D}_{\text{com}}-\hat{\mathbf{D}}_{\text{com}})  \big\|_2 \\
     &\mathcal{L}_{\text{str}}= \big\| \mathbf{M}_{\text{detail}}\odot (\mathbf{D}_{\text{str}}-\hat{\mathbf{D}}_{\text{str}})  \big\|_2 \\
     &\mathcal{L}_{\text{detail}} = \mathcal{L}_{\text{sta}} + \mathcal{L}_{\text{com}} + \mathcal{L}_{\text{str}}
\end{aligned},
    \label{Eq.disp}
\end{equation}
where $\mathbf{M}_{\text{detail}}$ is the facial mask in the UV coordinates, and $\hat{\mathbf{D}}_{\text{sta}}/\hat{\mathbf{D}}_{\text{com}}/\hat{\mathbf{D}}_{\text{str}}$ and $\mathbf{D}_{\text{sta}}/\mathbf{D}_{\text{com}}/\mathbf{D}_{\text{str}}$ are the reconstructed and ground-truth displacement maps, respectively.

\myparagraph{Coarse Shape Losses.}
Since the details should be based on realistic coarse shapes, we train the coarse shape to help the learning of details by leveraging the ground-truth vertex as supervision:
\begin{equation}
     \mathcal{L}_{\text{ver}} = \big\| \mathbf{M}_{\text{ver}}\odot (\mathbf{S}-\hat{\mathbf{S}})  \big\|_2,
    \label{Eq.vertexloss}
\end{equation}
where $\mathbf{M}_{\text{ver}}$ is frontal face area of the coarse shape. $\hat{\mathbf{S}}$ and $\mathbf{S}$ are the reconstructed and ground-truth face by Eq.~\ref{Eq.shape3dmm}.

In addition, we make constraints on shape coefficients to prevent overfitting. We enforce the predicted coefficients have a similar distribution to the ground-truth coefficients:
\begin{equation}
    \mathcal{L}_{\text{kl}}=\rho(\boldsymbol{\beta})\big(\log\rho(\boldsymbol{\beta})-\log\rho(\hat{\boldsymbol{\beta}})\big),
    \label{Eq.id_dist}
\end{equation}
where $\rho$ denotes \textit{softmax} function to map the predicted coefficients $\hat{\boldsymbol{\beta}}$ and ground-truth $\boldsymbol{\beta}$ into probability distribution.

Finally, the shape loss is $\mathcal{L}_{\text{shp}} = \mathcal{L}_{\text{ver}} + \mathcal{L}_{\text{kl}}$.

\myparagraph{Self-supervised Losses.}
To encourage the generalization of our models in real-world images~\cite{CelebAMask-HQ,mollahosseini2017affectnet},
we follow previous methods~\cite{feng2021learning,deng2019accurate,wood2022dense} to leverage self-supervised loss $\mathcal{L}_{\text{self}}$ for all training images, including photo loss $\mathcal{L}_{\text{pho}}$, perceptual loss $\mathcal{L}_{\text{id}}$, and dense landmark loss $\mathcal{L}_{\text{lmk}}$:
\begin{equation}
     \mathcal{L}_{\text{self}} = \mathcal{L}_{\text{pho}} + \lambda_{\text{id}}\mathcal{L}_{\text{id}} + \lambda_{\text{lmk}}\mathcal{L}_{\text{lmk}},
    \label{Eq.photo}
\end{equation}
where $\lambda_{\text{id}}$ and $\lambda_{\text{lmk}}$ are weights to balance the self-supervised losses term.

\begin{table*}[t!]
\caption{\textbf{Quantitative comparison of 3D face reconstruction methods on REALY benchmark}. ``-c'' and ``-d'' indicate coarse and detail shape, respectively. @$\mathcal{R}_N$/@$\mathcal{R}_M$/@$\mathcal{R}_F$/@$\mathcal{R}_C$/all indicate errors in nose/mouth/forehead/cheek/all regions. We highlight the best method for the two groups respectively. {\name} achieves the best reconstruction performance in the overall error by a large margin. Each component in {\name} contributes to a better reconstruction quality. The reconstructed details of {\name} further boost the quality while previous methods~\cite{feng2021learning,danvevcek2022emoca} modeling details with only image-level supervision even deteriorate the reconstruction accuracy.} \label{tab:benchmark}
{
\resizebox{1\linewidth}{!}{%
\begin{tabular}{r|r|ccccc|ccccc}
\toprule[1pt]
\multirow{2}{*}{Group}   & \multirow{2}{*}{\begin{tabular}[c]{@{}l@{}}Methods /\\ $e$ (mm)\end{tabular}}           & \multicolumn{5}{c|}{\textbf{\cellcolor{pink!20}  \textit{frontal-view}}}                                                                                 & \multicolumn{5}{c}{\cellcolor{gray!10} \textbf{\textit{side-view}}}                                                                                              \\
& &\multicolumn{1}{c}{\cellcolor{col_nose!30}@$\mathcal{R}_N$}  &\multicolumn{1}{c}{\cellcolor{col_mouth!30}@$\mathcal{R}_M$}   & \multicolumn{1}{c}{\cellcolor{col_forehead!30}@$\mathcal{R}_{F}$}   &                \multicolumn{1}{c}{\cellcolor{col_cheek!30} @$\mathcal{R}_{C}$}          &  \cellcolor{gray!30} all        &\multicolumn{1}{c}{\cellcolor{col_nose!30}@$\mathcal{R}_N$}  &\multicolumn{1}{c}{\cellcolor{col_mouth!30}@$\mathcal{R}_M$}   & \multicolumn{1}{c}{\cellcolor{col_forehead!30}@$\mathcal{R}_{F}$}   &                \multicolumn{1}{c}{\cellcolor{col_cheek!30} @$\mathcal{R}_{C}$}          &  \cellcolor{gray!30} all                         \\ \midrule[1pt]
\multirow{9}{*}{\begin{tabular}[c]{@{}c@{}@{}c@{}@{}}\rotatebox{0}{Coarse}\end{tabular}}
& Deep3D~\cite{deng2019accurate}                           & 1.719$\pm$0.354       & \textbf{1.368$\pm$0.439}       & 2.015$\pm$0.449       & \multicolumn{1}{c|}{1.528$\pm$0.501} & 1.657                 & 1.749$\pm$0.343       & 1.411$\pm$0.395      & 2.074$\pm$0.486       & \multicolumn{1}{c|}{1.528$\pm$0.517} & 1.691                \\
& MGCNet~\cite{shang2020self}                           & 1.771$\pm$0.380       & 1.417$\pm$0.409       & 2.268$\pm$0.503       & \multicolumn{1}{c|}{1.639$\pm$0.650} & 1.774                 & 1.827$\pm$0.383       & \textbf{1.409$\pm$0.418}       & 2.248$\pm$0.508       & \multicolumn{1}{c|}{1.665$\pm$0.644} & 1.787                \\
& 3DDFA-v2~\cite{guo2020towards}                         & 1.903$\pm$0.517       & 1.597$\pm$0.478       & 2.447$\pm$0.647       & \multicolumn{1}{c|}{1.757$\pm$0.642} & 1.926                 & 1.883$\pm$0.499       & 1.642$\pm$0.501       & 2.465$\pm$0.622       & \multicolumn{1}{c|}{1.781$\pm$0.636} & 1.943                \\
& DECA-c~\cite{feng2021learning}                           & 1.694$\pm$0.355       & 2.516$\pm$0.839       & 2.394$\pm$0.576       & \multicolumn{1}{c|}{1.479$\pm$0.535} & 2.010                 & 1.903$\pm$1.050       & 2.472$\pm$1.079       & 2.423$\pm$0.720       & \multicolumn{1}{c|}{1.630$\pm$1.135} & 2.107                \\
& SADRNet~\cite{ruan2021sadrnet}                          & 1.791$\pm$0.542       & 1.591$\pm$0.488       & 2.413$\pm$0.537       & \multicolumn{1}{c|}{1.856$\pm$0.701} & 1.913                 & 1.771$\pm$0.521       & 1.560$\pm$0.462       & 2.490$\pm$0.566       & \multicolumn{1}{c|}{2.010$\pm$0.715} & 1.958                \\
& EMOCA-c~\cite{danvevcek2022emoca}                          &    1.868$\pm$0.387                  & 2.679$\pm$1.112                      &       2.426$\pm$0.641               & \multicolumn{1}{c|}{1.438$\pm$0.501}               &            2.103           &   1.867$\pm$0.554                  &     2.636$\pm$1.284                 &   2.448$\pm$0.708                    &  \multicolumn{1}{c|}{1.548$\pm$0.590}               &   2.125                    \\

& MICA~\cite{MICA}                             & 1.585$\pm$0.325       & 3.478$\pm$1.204       & 2.374$\pm$0.683       & \multicolumn{1}{c|}{\textbf{1.099$\pm$0.324}} & 2.134                 & 1.525$\pm$0.322       & 3.567$\pm$1.212       & 2.379$\pm$0.675       & \multicolumn{1}{c|}{\textbf{1.109$\pm$0.325}} & 2.145                \\ 
& Ours-c (w/o Syn. Data)$^\dagger$               &     1.227$\pm$0.407            &    1.787$\pm$0.439            &    1.454$\pm$0.382           &             \multicolumn{1}{c|}{1.762$\pm$0.436}   &   1.558    &   1.187$\pm$0.379             &    1.826$\pm$0.490            &       1.470$\pm$0.426        &      \multicolumn{1}{c|}{1.653$\pm$0.450}          & 1.534   \\ 

& \textbf{Ours-c}             &   \textbf{1.054$\pm$0.317}            &        1.461$\pm$0.430        &   \textbf{1.331$\pm$0.347}            & \multicolumn{1}{c|}{1.342$\pm$0.384}                &   \multicolumn{1}{c|}{\textbf{1.297}}    &      \textbf{0.992$\pm$0.246}          &     1.505$\pm$0.454           &   \textbf{1.427$\pm$0.400}            & \multicolumn{1}{c|}{1.439$\pm$0.429}               &    \textbf{1.341}   \\ \midrule[1pt]
\multirow{7}{*}{\begin{tabular}[c]{@{}c@{}@{}c@{}@{}}\rotatebox{0}{Detail}\end{tabular}}
& DECA-d~\cite{feng2021learning}                           & 2.138$\pm$0.461       & 2.802$\pm$0.868       & 2.457$\pm$0.559       & \multicolumn{1}{c|}{1.443$\pm$0.498} & 2.210                 & 2.286$\pm$1.103       & 2.684$\pm$1.041       & 2.519$\pm$0.718       & \multicolumn{1}{c|}{1.555$\pm$0.822} & 2.261                \\
& EMOCA-d~\cite{danvevcek2022emoca}                          &   2.532$\pm$0.539                   &     2.929$\pm$1.106                 &  2.595$\pm$0.631                    &  \multicolumn{1}{c|}{1.495$\pm$0.469}               &  2.388              &   2.455$\pm$0.636                   &     2.948$\pm$1.292                 &  2.606$\pm$0.686                    &  \multicolumn{1}{c|}{1.599$\pm$0.563}               &  2.402              \\
& HRN~\cite{lei2023Ahierarchical}                             & 1.722$\pm$0.330       & \textbf{1.357$\pm$0.523}       & 1.995$\pm$0.476      & \multicolumn{1}{c|}{\textbf{1.072$\pm$0.333}} & 1.537                 & 1.642$\pm$0.310      & \textbf{1.285$\pm$0.528}       & 1.906$\pm$0.479       & \multicolumn{1}{c|}{\textbf{1.038$\pm$0.322}} & 1.468               \\ 
& Ours-d (w/o Syn. Data)$^\dagger$                &    1.465$\pm$0.557            &  1.790$\pm$0.425              &      1.528$\pm$0.373         &            \multicolumn{1}{c|}{1.618$\pm$0.362}    &    1.600    &    1.422$\pm$0.537            &   1.849$\pm$0.473             &   1.530$\pm$0.414            &      \multicolumn{1}{c|}{1.572$\pm$0.399}          &   1.594  \\
& Ours-d (w/o static)$^*$                &       1.055$\pm$0.290         &  1.469$\pm$0.415              &     1.336$\pm$0.337          &    \multicolumn{1}{c|}{1.319$\pm$0.374}    &     1.295  &    1.004$\pm$0.233                &    1.491$\pm$0.437           &   1.418$\pm$0.392            &       \multicolumn{1}{c|}{1.418$\pm$0.415}         &   1.332 \\
& Ours-d (w/o dynamic)$^*$                &    1.069$\pm$0.318            &    1.469$\pm$0.414            &         1.358$\pm$0.336      &             \multicolumn{1}{c|}{1.270$\pm$0.344}   &   1.292    &       0.991$\pm$0.239         &    1.496$\pm$0.437            &      1.411$\pm$0.393         &      \multicolumn{1}{c|}{1.375$\pm$0.402}          & 1.318   \\
& \textbf{Ours-d}               &    \textbf{1.036$\pm$0.280}            &   1.450$\pm$0.413             &    \textbf{1.324$\pm$0.334}           & \multicolumn{1}{c|}{1.291$\pm$0.362}               &   \textbf{1.275}    &     \textbf{0.985$\pm$0.237}           &    1.489$\pm$0.436            &     \textbf{1.399$\pm$0.388}           & \multicolumn{1}{c|}{1.360$\pm$0.395}               &   \textbf{1.308} \\
\bottomrule[1pt]
\end{tabular}
}
\begin{tablenotes}
\scriptsize
\item $^\dagger$ To align the dataset scale, w/o Syn. Data indicates we train the model without using the ground-truth labels from the synthetic dataset.
\item $^*$ To eliminate the bias of coarse shape in estimating the reconstruction error, we fix the coarse shape and train the details with/without static and dynamic factors for comparisons.
\end{tablenotes}
}
\vspace{-5pt}
\end{table*}

In addition, considering the static detail heavily correlates to person-specific age attribute, inspired by~\cite{danvevcek2022emoca}, we leverage the pre-trained age prediction network~\cite{karkkainenfairface} to learn high-level representations of static details through knowledge distillation, such that the learned coefficients exhibit expressive results.
To achieve this, we use several MLP layers on the static coefficient $\boldsymbol\varphi$, and map it into age classification probabilities $\hat{\mathbf{p}}_{\text{age}}$. Then we use the pre-trained age recognition model $\mathbf{\Gamma}_{\text{age}}$ to obtain the probabilities of the given input image $\mathbf{I}$. The distillation loss $\mathcal{L}_{\text{kd}}$ enforces the probabilities between $\hat{\mathbf{p}}_{\text{age}}$ and $\mathbf{\Gamma}_{\text{age}}(\mathbf{I})$ be similar:
\begin{equation}
     \mathcal{L}_{\text{kd}} =\mathbf{\Gamma}_{\text{age}}(\mathbf{I})\big(\log\mathbf{\Gamma}_{\text{age}}(\mathbf{I})-\log\hat{\mathbf{p}}_{\text{age}}\big).
    \label{Eq.logit_distil}
\end{equation}
\myparagraph{Regularization.}
$\mathcal{L}_{\text{reg}}$ regularizes coefficients of each sub-module, by minimizing the $l_2$ loss of $\boldsymbol\alpha$, $\boldsymbol\beta$, $\boldsymbol\xi$, $\boldsymbol\varphi$, $\boldsymbol\phi$.

\myparagraph{Overall Loss Function.}
We train the coarse shape and fine details simultaneously, such that each component can collaborate to reconstruct high-fidelity 3D faces with realistic details. Formally, we minimize the total loss function:
\begin{equation}
\begin{aligned}
    \mathcal{L}&= \lambda_{\text{detail}}\mathcal{L}_{\text{detail}} + \lambda_{\text{shp}}\mathcal{L}_{\text{shp}}  \\ &+  \lambda_{\text{self}}\mathcal{L}_{\text{self}} +
    \lambda_{\text{kd}}\mathcal{L}_{\text{kd}} + \lambda_{\text{reg}}\mathcal{L}_{\text{reg}},    
\end{aligned}
\end{equation}
where $\lambda$ is the weight for each component.

\section{Experiments}

\subsection{Implementation Details}

\myparagraph{Dataset.}
We use a hybrid dataset made up from both synthetic~\cite{wood2021fake,raman2022mesh} and real data~\cite{CelebAMask-HQ,mollahosseini2017affectnet}. 
We use the synthetic data pipeline~\cite{wood2021fake,raman2022mesh} to generate a diverse dataset of $200k$ faces with ground-truth vertex, landmark, albedo, and displacement map annotations.
The real-world datasets contain $400k$ images in total from diverse age, gender, and ethnicity groups.
For the real-world dataset, we use the pre-trained dense landmark detector~\cite{wood2021fake} to detect $669$ landmarks for training.
We use face parsing~\cite{Zheng2022DecoupledML} to generate and select region-of-interest as facial masks, providing robustness to common occlusions by hair or other accessories.
We follow~\cite{danvevcek2022emoca,wood2022dense,deng2019accurate} to split the dataset into training and validation sets. The test images are from CelebA~\cite{CelebAMask-HQ}, FFHQ~\cite{karras2019style}, LS3D-W~\cite{bulat2017far}, and AFLW2000~\cite{yin2017towards}.

\begin{figure*}[t!]
\begin{minipage}[t]{0.48\linewidth}
    \begin{overpic}[trim=0cm 0cm 0cm 0cm,clip,width=1\linewidth,grid=false]{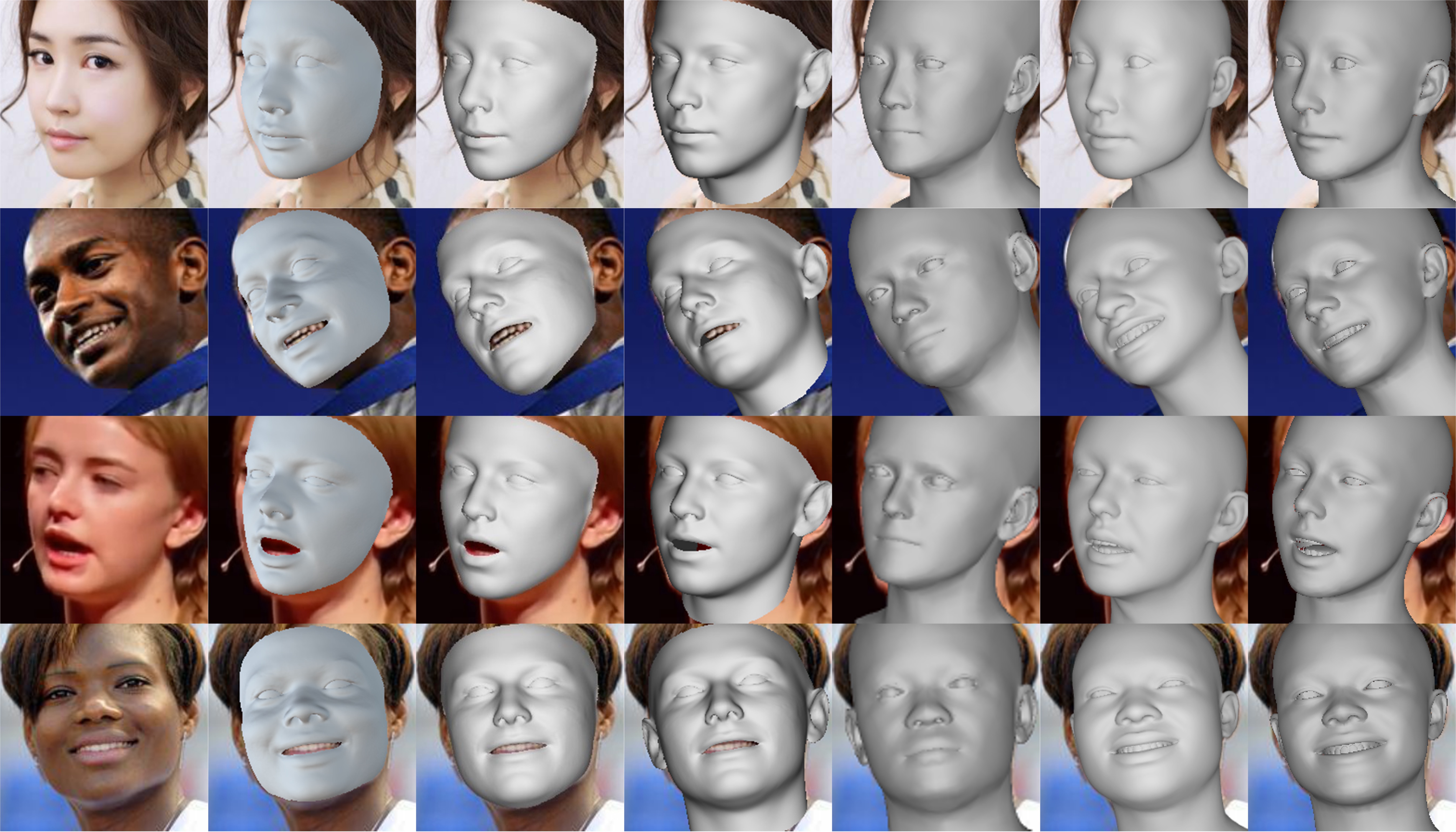}
    \end{overpic}
    \put(-228,138){\bfseries\scriptsize Input}
    \put(-25,138){\bfseries\scriptsize Ours}
    \put(-196,138){\scriptsize Deep3D}
    \put(-168,138){\scriptsize 3DDFA-v2}
    \put(-135,138){\scriptsize SynergyNet} 
    \put(-93,138){\scriptsize MICA} 
    \put(-60,138){\scriptsize Dense}
    \vspace{-5pt}
    \caption{\textbf{Comparison on coarse shape reconstruction.} From left to right: Input image, Deep3D~\cite{deng2019accurate}, 3DDFA-v2~\cite{guo2020towards}, SynergyNet~\cite{wu2021synergy}, MICA~\cite{MICA}, Dense~\cite{wood2022dense}, and {\name} (Ours).}
    \label{fig:coarse_cmp}
\end{minipage}%
    \hfill%
\begin{minipage}[t]{0.48\linewidth}
    \begin{overpic}[trim=0cm 0cm 0cm 0cm,clip,width=1\linewidth,grid=false]{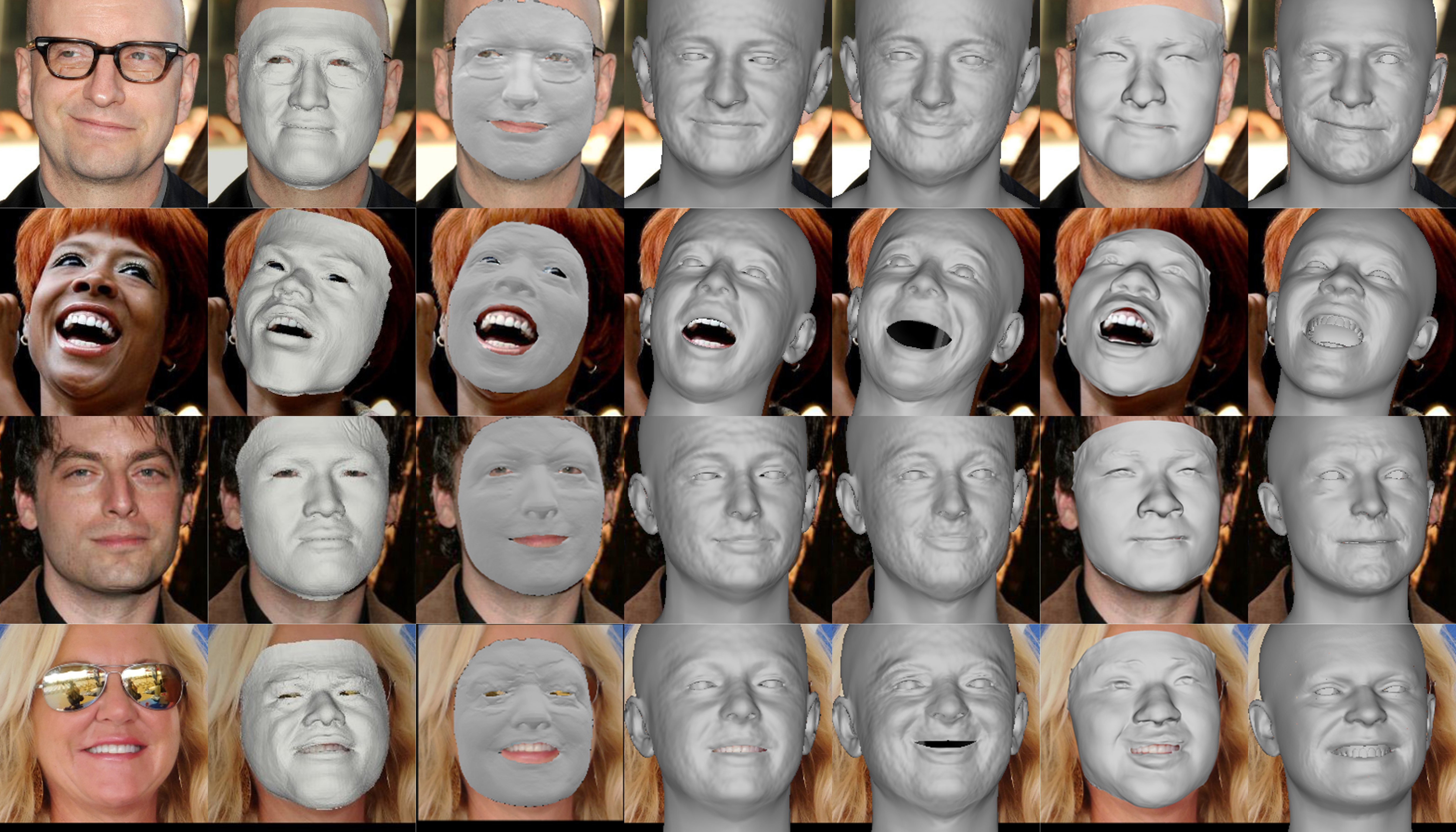}
    \end{overpic}
    \put(-228,138){\bfseries\scriptsize Input}
    \put(-25,138){\bfseries\scriptsize Ours}
    \put(-200,138){\scriptsize FaceScape}
    \put(-160,138){\scriptsize Unsup}
    \put(-128,138){\scriptsize DECA} 
    \put(-98,138){\scriptsize EMOCA} 
    \put(-65,138){\scriptsize FaceVerse}
    \vspace{-5pt}
    \caption{\textbf{Comparison on detail shape reconstruction.} From left to right: Input image, FaceScape~\cite{yang2020facescape}, Unsup~\cite{chen2020self}, DECA~\cite{feng2021learning}, EMOCA~\cite{danvevcek2022emoca}, FaceVerse~\cite{wang2022faceverse}, and {\name} (Ours).}
    \label{fig:detail_cmp}
\end{minipage} 
\vspace{-10pt}
\end{figure*}

\begin{figure}[!t]
    \centering
    \begin{overpic}[trim=0cm 0cm 0cm 0cm,clip,width=1\linewidth,grid=false]{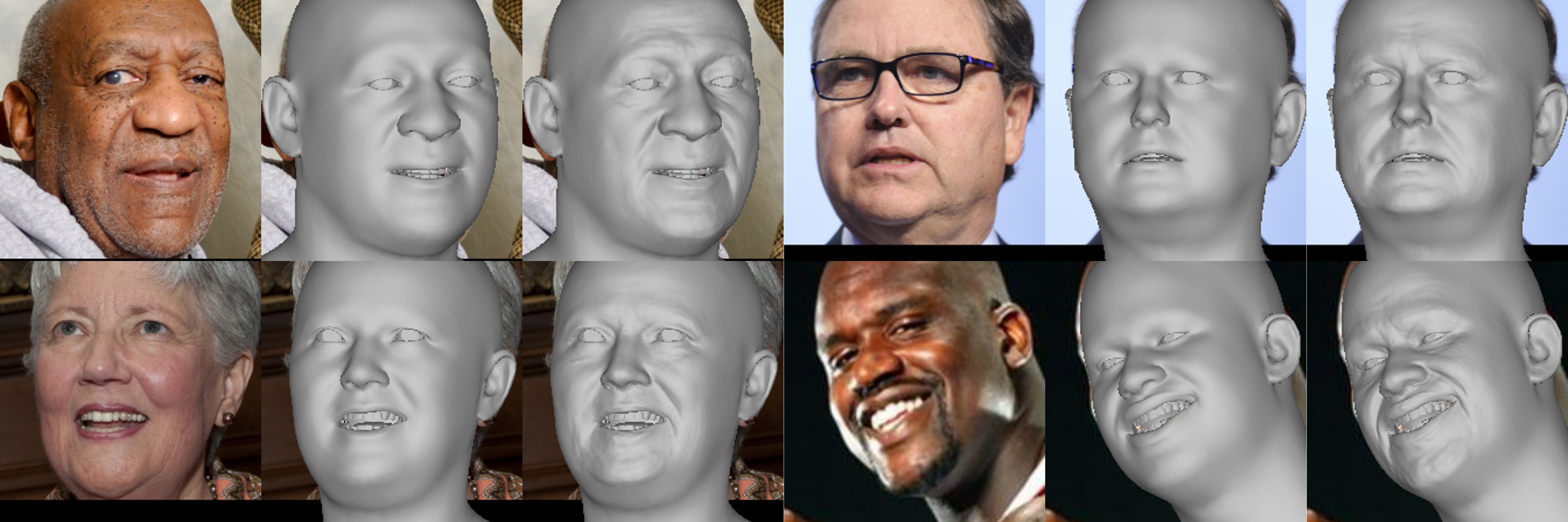}
    \end{overpic}
    \put(-225,81){\bfseries\scriptsize Input}
    \put(-195,81){\scriptsize Dense~\cite{wood2022dense}}  
    \put(-158,81){\bfseries\scriptsize + Our Detail}   
    \put(-105,81){\bfseries\scriptsize Input}
    \put(-75,81){\scriptsize  Dense~\cite{wood2022dense}}  
    \put(-38,81){\bfseries\scriptsize + Our Detail}  
    \vspace{-5pt}
    \caption{\textbf{Illustration on the flexibility of {\module}.} Given the identity and expression coefficients ($\boldsymbol{\beta}$, $\boldsymbol{\xi}$) from the optimization-based method~\cite{wood2022dense}, {\module} can generate realistic details based on the coarse shape and further improve the visual quality.}
    \label{fig:ft_detail}
    \vspace{-10pt}
\end{figure}
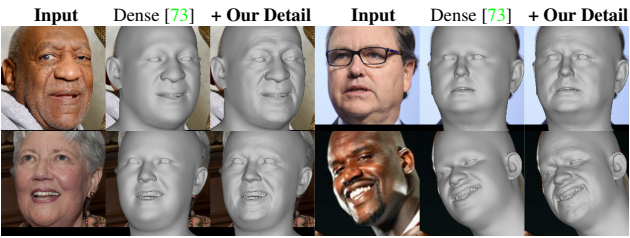

\myparagraph{Implementation Details.}
We implement {\name} in PyTorch~\cite{paszke2019pytorch} and leverage the PyTorch3D differentiable rasterizer~\cite{ravi2020pytorch3d} for rendering. 
We train our model for $35$ epochs on $8\times$ NVIDIA Tesla V100 GPUs with a mini-batch of $320$. We use the pre-trained ResNet-50 on ImageNet~\cite{deng2009imagenet} as initialization, and use Adam~\cite{kingma2015adam} as optimizer with an initial learning rate of $1e-4$. The input image is cropped and aligned by~\cite{chen2016supervised}, and resized into $224\times224$.
We empirically set $\lambda_{\text{detail}}=10$, $\lambda_{\text{shp}}=1$, $\lambda_{\text{self}}=1$, $\lambda_{\text{id}}=0.1$, $\lambda_{\text{lmk}}=0.5$, $\lambda_{\text{kd}}=1$, $\lambda_{\text{reg}}=1e-3$ throughout the experiments.

\subsection{Comparisons to State-of-the-art}

\myparagraph{Quantitative Comparison.}
We perform the quantitative evaluation on the REALY benchmark~\cite{REALY}, which contains $100$ frontal-view and $400$ side-view images from $100$ textured-scans~\cite{LYHM}. The REALY benchmark presents a region-aware evaluation pipeline to separately evaluate the metric error (in mm) of the nose, mouth, forehead, and cheek regions. Such an evaluation pipeline is demonstrated to better estimate the actual similarity of the 3D faces and align with human perception.
We compare {\name} to previous state-of-the-art methods and report the region-wise and average normalized mean square error (NMSE) in Tab.~\ref{tab:benchmark}.

As Tab.~\ref{tab:benchmark} illustrates, {\name} outperforms prior arts in the overall error by a large margin. {\name} balances the reconstruction quality on each region, compared to those optimum region methods that may fail in specific regions ({\eg}, MICA~\cite{MICA} fails in mouth region while HRN~\cite{lei2023Ahierarchical} fails in forehead region).
Note that {\name} faithfully recovers the facial details, thus making the reconstruction error smaller than using the coarse shape alone. As a comparison, although DECA~\cite{feng2021learning} and EMOCA~\cite{danvevcek2022emoca} can reconstruct details of given images, they turn out to be noisy, leading to the deterioration of reconstruction quality.

In addition, Tab.~\ref{tab:benchmark} also demonstrates the necessity of each component in contributing to a better quality.
It can be observed that the synthetic data with ground-truth labels not only improve the coarse shape reconstruction quality but is also crucial for detailed reconstruction.
With the synthetic data, the proposed {\module} further boosts the overall reconstruction quality. Both the static and dynamic factors are essential to capture fine-grained details, and the final {\module} achieves the most accurate details in expression-related regions such as the mouth and forehead, which contributes to the overall gains.

\myparagraph{Qualitative Comparison.}
Given a single face image, {\name} reconstructs a high-fidelity 3D shape with details.
We present comparisons with previous methods on 1). coarse shape reconstruction~\cite{deng2019accurate,guo2020towards,wu2021synergy,MICA,wood2022dense} in Fig.~\ref{fig:coarse_cmp} and 2). detail reconstruction~\cite{yang2020facescape,chen2020self,feng2021learning,danvevcek2022emoca,wang2022faceverse} in Fig.~\ref{fig:detail_cmp}. See more examples and comparisons in the supplementary.

For coarse shape in Fig.~\ref{fig:coarse_cmp}, our {\name} faithfully recovers the coarse shape of the given identity and outperforms the previous learning-based methods, and is on par with Dense~\cite{wood2022dense}, which is the state-of-the-art optimization-based method.
For detailed reconstruction in Fig.~\ref{fig:detail_cmp}, previous methods~\cite{feng2021learning,danvevcek2022emoca} fail to reconstruct satisfactory details. Several methods~\cite{chen2020self,yang2020facescape,wang2022faceverse} are sensitive to occlusions and large poses. 
As a comparison, {\name} achieves the most realistic reconstruction quality, and faithfully recovers facial details of a given image, which significantly outperforms previous methods by a large margin.

\begin{figure}[!t]
    \centering
    \begin{overpic}[trim=0cm 0cm 0cm 0cm,clip,width=1\linewidth,grid=false]{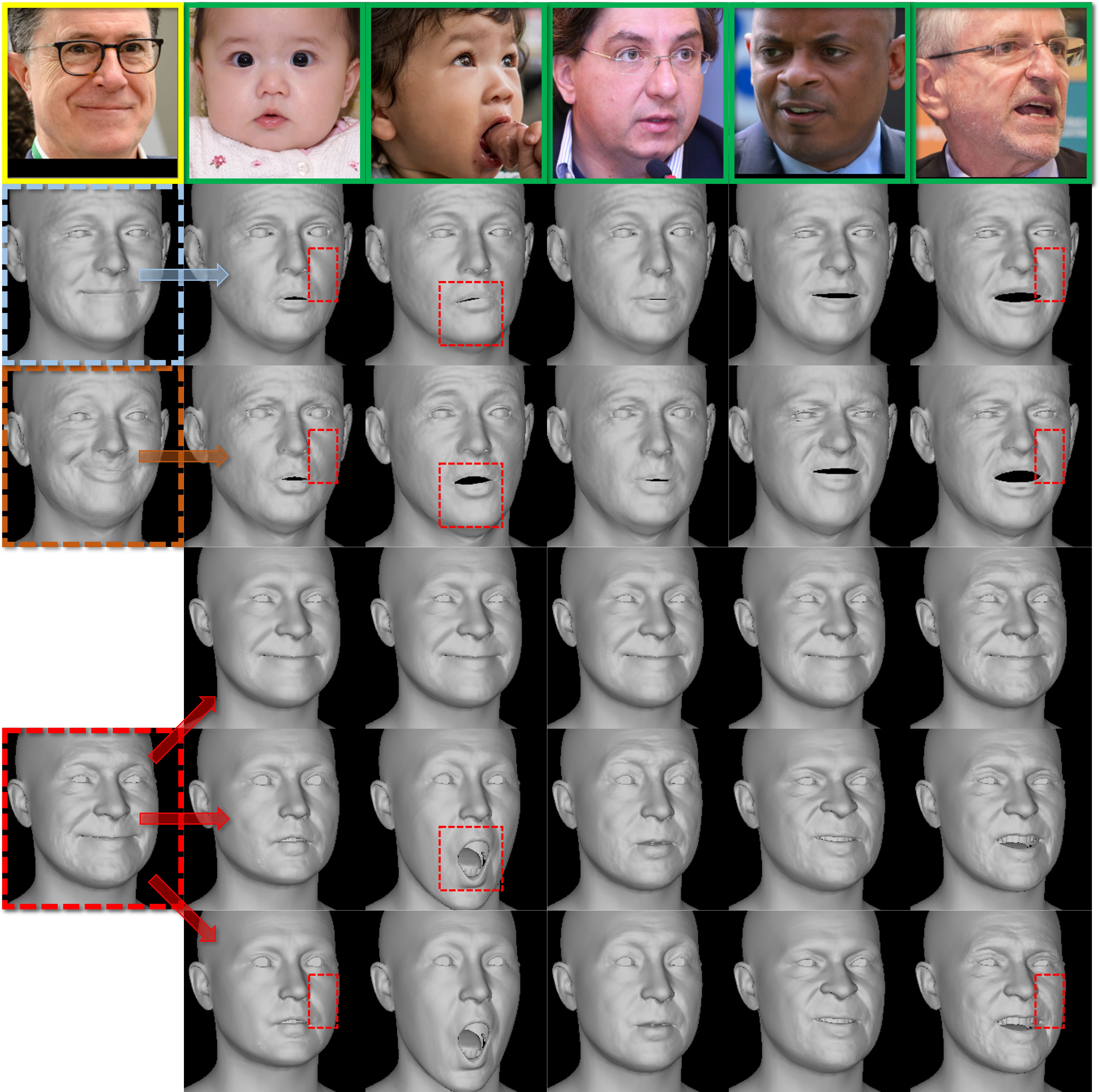}
    \end{overpic}
    \put(-225,30){\bfseries\scriptsize Ours}    
    \vspace{-5pt}
    \caption{\textbf{Comparison on face animation.} Given a source image (yellow box), we use the driving images (green box) to drive its expressions. DECA~\cite{feng2021learning} (2nd-row) and EMOCA~\cite{danvevcek2022emoca} (3rd-row) can animate the expression-driven details but lack realistic. As a comparison, {\name} is flexible to animate details from static (4th-row), dynamic (5th-row), or both (6th-row) factors, and presents vivid animation quality with realistic shapes.
    }
    \label{fig:animation}
    \vspace{-10pt}
\end{figure}

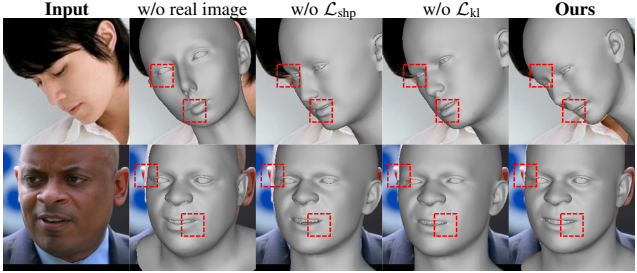
\begin{figure*}[t!]
\begin{minipage}[t]{0.48\linewidth}
    \begin{overpic}[trim=0cm 0cm 0cm 0cm,clip,width=1\linewidth,grid=false]{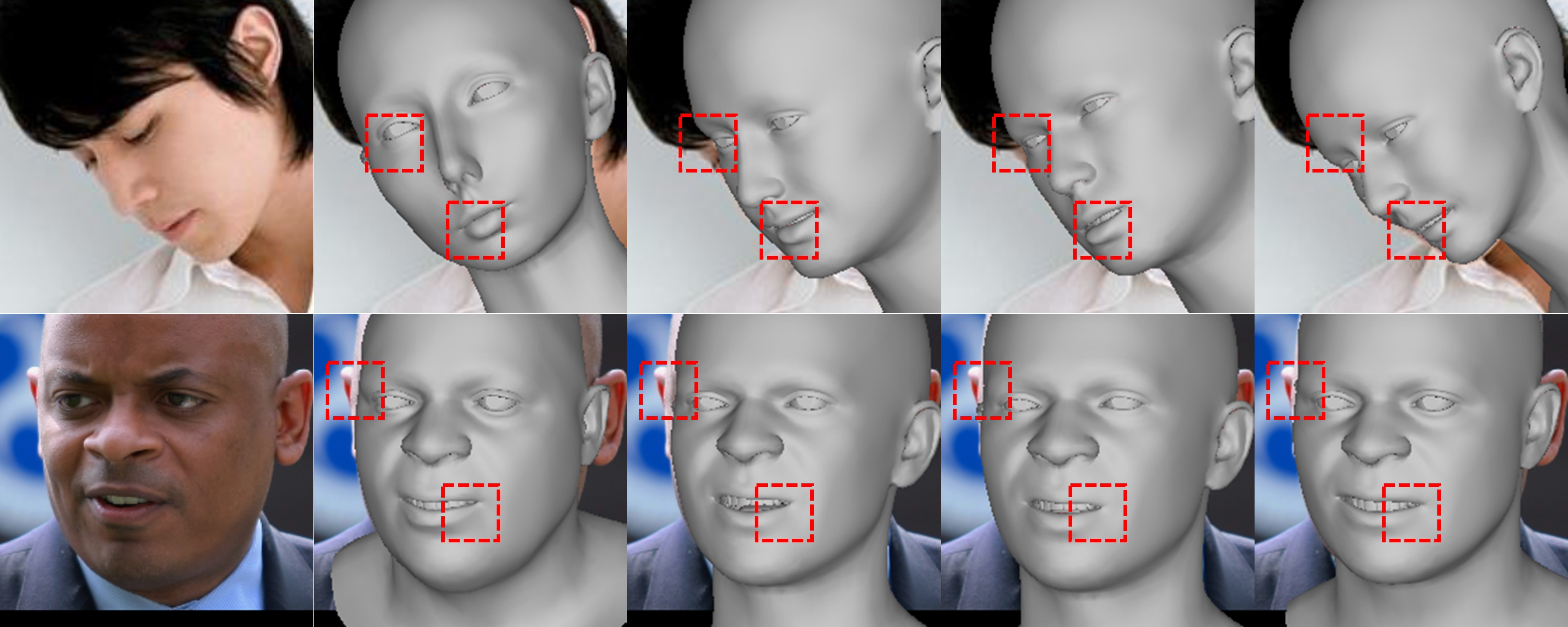}
    \end{overpic}
    \put(-223,97){\bfseries\scriptsize Input}
    \put(-188,97){\scriptsize w/o real image}
    \put(-130,97){\scriptsize w/o $\mathcal{L}_{\text{shp}}$}
    \put(-80,97){\scriptsize w/o $\mathcal{L}_{\text{kl}}$}
    \put(-30,97){\bfseries\scriptsize Ours}
    \label{fig:coarse_loss}
\end{minipage}%
    \hfill%
\begin{minipage}[t]{0.48\linewidth}
    \begin{overpic}[trim=0cm 0cm 0cm 0cm,clip,width=1\linewidth,grid=false]{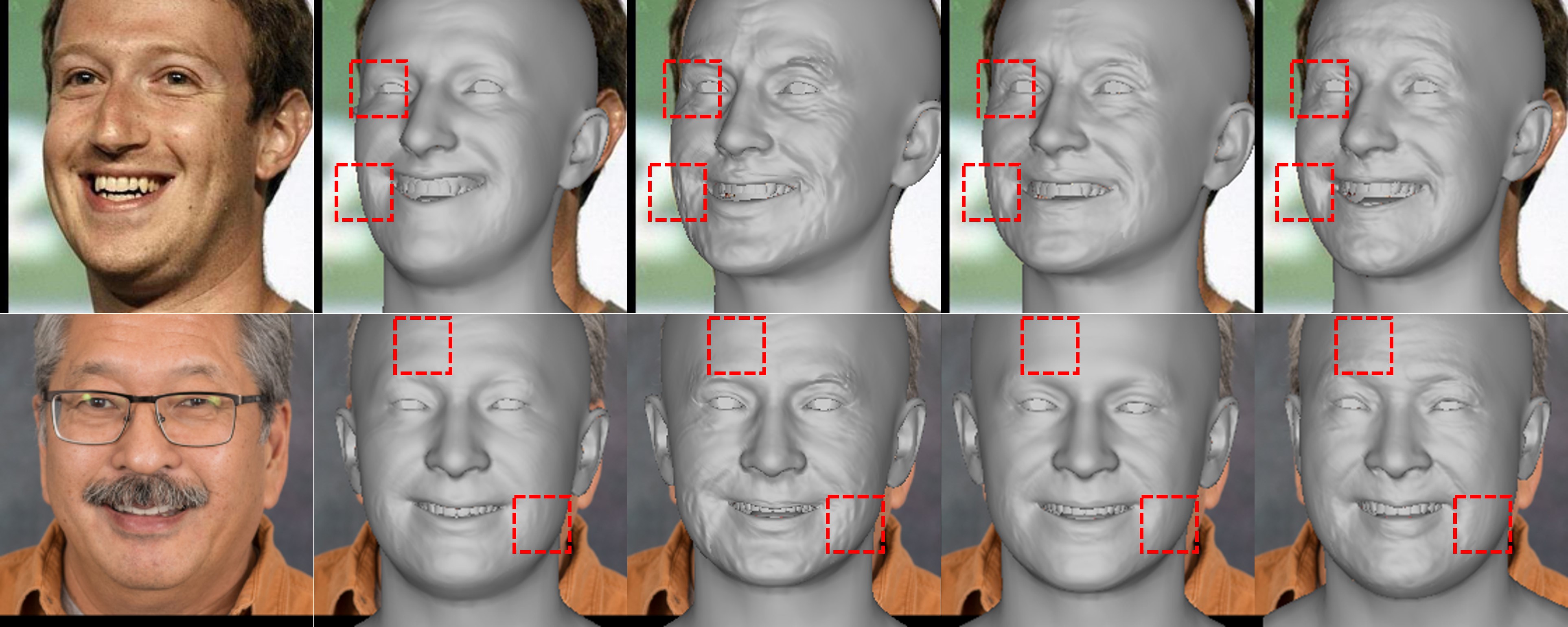}
    \end{overpic}
    \put(-223,97){\bfseries\scriptsize Input}
    \put(-188,97){\scriptsize w/o real image}
    \put(-132,97){\scriptsize w/o $\mathcal{L}_{\text{detail}}$}
    \put(-80,97){\scriptsize w/o $\mathcal{L}_{\text{kd}}$}
    \put(-30,97){\bfseries\scriptsize Ours} 
    \label{fig:detail_loss}
\end{minipage}
\vspace{5pt}
\caption{\textbf{Ablation studies on loss functions and training data.} The coarse shape losses $\mathcal{L}_{\text{shp}}$/$\mathcal{L}_{\text{kl}}$ (left), detail losses $\mathcal{L}_{\text{detail}}$/$\mathcal{L}_{\text{kd}}$ (right), and hybrid datasets coherently contribute to the reconstruction quality of coarse shapes and details.}
\label{fig.abl_loss}
\vspace{-10pt}
\end{figure*}

In addition, given an image and the fitted coefficients $\boldsymbol{\beta}$, $\boldsymbol{\xi}$ from the optimization-based methods such as Dense~\cite{wood2022dense}, {\module} synthesizes the details and further strengthens the quality compared to the coarse shape (see Fig.~\ref{fig:ft_detail}). It shows {\module} is flexible and can be easily plugged-and-play into other methods. 
See more in the supplementary.

\myparagraph{Application of {\name}.}
The {\name} explicitly decouples the static and dynamic details through the proposed {\module}.
Therefore, we can animate the facial attributes by simply assigning the expression coefficient $\boldsymbol{\xi}$ and/or static coefficient $\boldsymbol{\varphi}$ of the driving images to the source images.

In Fig.~\ref{fig:animation}, we demonstrate the animation quality of {\name} outperforms the previous state-of-the-art detail animation methods~\cite{feng2021learning,danvevcek2022emoca}.
It shows that while DECA~\cite{feng2021learning} and EMOCA~\cite{danvevcek2022emoca} can animate the expression-driven details but lack realistic, the proposed {\name} is flexible to manipulate the static, dynamic, or both details.
When animating the static detail, the person-specific details can be well transferred into the source shape.
When animating the dynamic detail, only expression-dependent details are presented.
Finally, we can also animate the static and dynamic details simultaneously and achieve satisfactory results.

\section{Ablation Studies}

\myparagraph{Ablation Studies on Loss Functions and Datasets.}
We present ablation studies on the proposed loss functions and training strategy with hybrid datasets.
We train {\name} with synthetic data alone and compare it to the one trained with hybrid datasets.
For coarse shape reconstruction, we investigate the contribution of $\mathcal{L}_{\text{shp}}$ and its sub-term $\mathcal{L}_{\text{kl}}$.
For detail reconstruction, we compare {\name} without $\mathcal{L}_{\text{detail}}$ and $\mathcal{L}_{\text{kd}}$, respectively. The results are presented in Fig.~\ref{fig.abl_loss}.

Fig.~\ref{fig.abl_loss} demonstrates that the proposed loss functions and training strategy from hybrid datasets contribute to satisfactory coarse shape and details.
First, the model trained with synthetic data alone cannot generalize well to real-world images, which indicates the necessity to train with real-world data.
Second, $\mathcal{L}_{\text{shp}}$ improves the coarse shape reconstruction quality. $\mathcal{L}_{\text{shp}}$ is effective in tackling challenging poses and improving alignment. $\mathcal{L}_{\text{kl}}$ can relieve the overfitting risk on the synthetic data and improve the generalization to real-world images.
Third, without $\mathcal{L}_{\text{detail}}$ or $\mathcal{L}_{\text{kd}}$, the reconstructed details exhibit random noise and cannot faithfully reflect person-specific details. Such noise misses the correspondence to the person-specific identity.

\begin{figure}[!t]
    \centering
    \begin{overpic}[trim=0cm 0cm 0cm 0cm,clip,width=1\linewidth,grid=false]{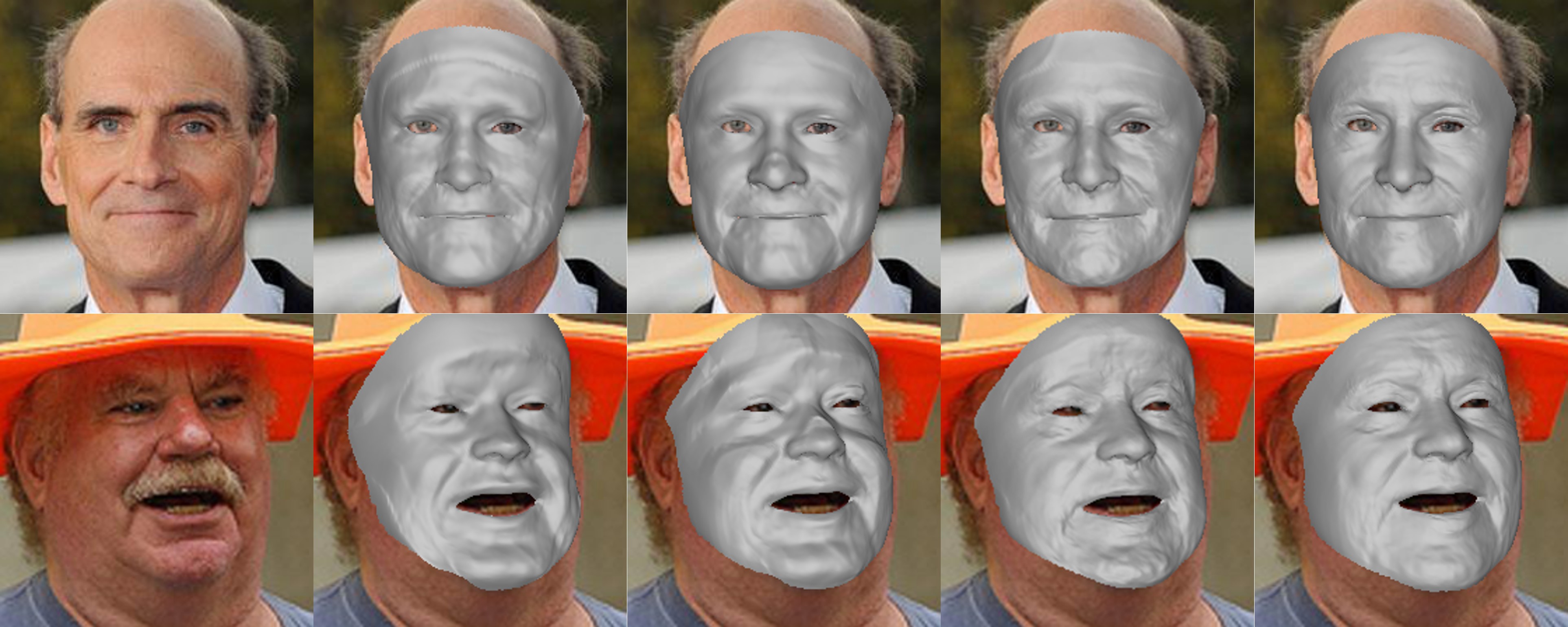}
    \end{overpic}
    \put(-223,97){\bfseries\scriptsize Input}
    \put(-172,97){\scriptsize SD-1}
    \put(-125,97){\scriptsize SD-2}
    \put(-78,97){\scriptsize SD-3}
    \put(-30,97){\bfseries\scriptsize Ours} 
    \vspace{-5pt}
    \caption{\textbf{Ablation studies on {\module}.}
    Results show that directly synthesizing the static or dynamic details is rather challenging, leading to unreasonable coarse shapes and details (SD-1, SD-2, and SD-3). 
    As a comparison, we leverage the statistical bases with {\module} and regard the detail generation problem as a coefficients regression and interpolation problem, leading to more realistic details.}
    \label{fig:abldynamic}
    \vspace{-10pt}
\end{figure}

\myparagraph{Ablation Studies on {\module}.}
To verify the effectiveness of building bases for static and dynamic details, we present detailed ablation studies on {\module}, by replacing the bases ({\ie}, $\mathbf{B}_{\text{sta}}$ and $\mathbf{B}_{\text{com}}/\mathbf{B}_{\text{str}}$) reconstruction with a U-Net decoder~\cite{ronneberger2015u} (same as DECA~\cite{feng2021learning}). Therefore, the model learns to directly synthesize displacement maps instead of predicting corresponding coefficients like ours.
In Fig.~\ref{fig:abldynamic}, we make comparisons on: 1). directly synthesizing $\mathbf{D}_{\text{dyn}}$ (SD-1), 2). directly synthesizing $\mathbf{D}_{\text{com}}/\mathbf{D}_{\text{str}}$ and interpolating via Eq.~\ref{Eq.dyn_detail} (SD-2), 3). directly synthesizing $\mathbf{D}_{\text{sta}}$ (SD-3), 4). {\module} (Ours).
It can be seen that, due to the high diversity and complexity of expression representation, it is hard to directly learn realistic details even with ground-truth labels of $\mathbf{D}_{\text{dyn}}$ from synthetic data (see SD-1, SD-2 and SD-3 in Fig.~\ref{fig:abldynamic}).
More specifically, for the static details, directly synthesizing displacement maps bring much noise (SD-3). For example, the hollow eyebrow is demonstrated in the second row.
For the dynamic details, directly synthesizing displacement maps even leads to unnatural results (SD-1 and SD-2). For example, the reconstructed 3D faces are distorted especially in the second row. We also notice that, directly synthesizing $\mathbf{D}_{\text{dyn}}$ (SD-1) achieves inferior results than directly synthesizing $\mathbf{D}_{\text{com}}/\mathbf{D}_{\text{str}}$ and interpolating via Eq.~\ref{Eq.dyn_detail} (SD-2). This demonstrates that it is beneficial to simplify the expression representation by using interpolation between two displacement maps (\ie, compressed and stretched).
In conclusion, these observations further verify our insight on relaxing the challenging detail generation problem into a feasible coefficients regression problem.

\section{Conclusion}
We propose {\name} to reconstruct high-fidelity 3D faces with realistic and animatable details from a single image. 
Our motivation and insights stand on the spirit of 3DMMs to simplify the challenging detail generation into more accessible regression and interpolation tasks.
To achieve this, we elaborately design {\module} to decouple the static and dynamic factors explicitly, and interpolate the dynamic details through vertex tension. We succeed in learning the coarse shape and details jointly by proposing several novel loss functions to train on synthetic and real-world data. 
Extensive experiments demonstrate that {\name} achieves state-of-the-art face reconstruction both quantitatively and qualitatively in the coarse shape and detail shape, and the details are well decoupled and naturally animatable.

\section*{Acknowledgement}
This work was partially supported by the National Key R\&D Program of China (2022YFB4701400/4701402), SZSTC Grant (JCYJ20190809172201639, WDZC2020-0820200655001), Shenzhen Key Laboratory (ZDSYS2021-0623092001004), and Beijing Key Lab of Networked Multimedia.


\balance{
{\small
\bibliographystyle{ieee_fullname}
\bibliography{main}
}
}

\newpage
\appendix
\renewcommand{\thetable}{A\arabic{table}}
\renewcommand{\thefigure}{A\arabic{figure}}
\renewcommand{\theequation}{A\arabic{equation}}
\twocolumn[
\begin{@twocolumnfalse}
{
\begin{center}
    \Large \textbf{\name: High-Fidelity 3D Face Reconstruction \\ by Learning Static and Dynamic Details }\\
    \Large \textbf{------------------------Appendix------------------------}
\end{center}
}
{
\vspace*{24pt}
\large
\lineskip .5em
\centering
\begin{tabular}[t]{c}
\bigskip
 \vspace*{1pt}\\%
\end{tabular}
\par
\vskip .5em
\vspace*{12pt}
}
\end{@twocolumnfalse}
]

In this supplementary, we provide additional results and discussions to make our paper self-contained. Specifically, we present 1). more details on experiments and implementation in Sec.~\ref{sec.supp_imp}, 2). details of loss functions in Sec.~\ref{sec.supp_loss}, 3). more experimental comparisons and results in Sec.~\ref{sec.supp_exp}, 4). discussions on limitations and future work in Sec.~\ref{sec.supp_limit}, and 5). discussions on social impact in Sec.~\ref{sec.supp_social}.

\section{Experimental \& Implementation Details} \label{sec.supp_imp}

\subsection{Details of Dataset}

We follow the synthetic data pipeline~\citesupp{raman2022meshsupp,wood2021fakesupp} to synthesize $200k$ images, consisting of $40k$ identities with $5$ frames. For each identity, we assign different expressions, viewpoints, illumination, accessories, and backgrounds to improve the model's robustness for training.

In addition, in the synthetic data pipeline, the ground-truth albedo, neutral displacement maps, stretched displacement maps, and compressed displacement maps are sampled from the $332$ captured scans~\citesupp{raman2022meshsupp,wood2021fakesupp} and recorded. 
In this paper, to ensure the efficiency of our model, we resize the $4096\times 4096$ resolution assets into $512\times 512$ as the ground-truth labels for training. 
For the real-world data~\citesupp{CelebAMask-HQsupp,mollahosseini2017affectnetsupp}, we split them into the train ($400k$) and valid ($30k$). During training, common data augmentation techniques ({\ie}, random shift, scale, rotation, and flip) are adopted to improve the robustness of our model.

\subsection{Details of Ablation Studies}

In the ablation on datasets, we only remove the real-world data while keeping other settings the same as our full {\name}. %
In the ablation on loss functions, we only remove specific loss functions while training the models that follow the same settings as our full {\name}.

For the ablation studies on {\module}, in SD-1, instead of obtaining the dynamic detail by interpolating the compressed and stretched displacement maps as in {\module}, we directly synthesize the dynamic detail by a learnable network end-to-end. The ground-truth labels for the synthetic dataset follow the dynamic composition in Eq.~\ref{Eq.dyn_detail}.
More specifically, we use an MLP layer to map expression-aware $\boldsymbol{\phi}$ into $128$-dim latent code $\mathbf{z_1}$, and follow~\citesupp{feng2021learningsupp} to use a U-Net decoder~\citesupp{ronneberger2015usupp} to synthesize corresponding dynamic displacement map in $512\times 512$ resolution. 

In SD-2, instead of generating compressed and stretched displacement maps using the PCA bases as in {\module}, we employ two learnable networks to synthesize the compressed and stretched displacement maps in parallel, and compound the dynamic details by interpolating the two displacement maps using Eq.~\ref{Eq.dyn_detail}. 
Similar to SD-1, we use two MLP layers to map $\boldsymbol{\phi}$ into $128$-dim latent codes $\mathbf{z_2}$ and $\mathbf{z'_2}$, and use two U-Net decoders to synthesize corresponding polarized displacement maps in $512\times 512$ resolution.

In SD-3, instead of generating the static displacement map using the PCA basis as in {\module}, we directly synthesize the static detail by a learnable network end-to-end. More specifically, we use an MLP layer to map age-aware $\boldsymbol{\varphi}$ into $128$-dim latent code $\mathbf{z_3}$, and follow~\citesupp{feng2021learningsupp} to use a U-Net decoder~\citesupp{ronneberger2015usupp} to synthesize corresponding static displacement map in $512\times 512$ resolution.

\section{Details of Loss functions} \label{sec.supp_loss}

In our main paper, we propose several simple yet effective loss functions to train our model end-to-end using both synthetic and real-world data. As we have justified through the detailed ablation studies, each loss function contributes to the coarse shape and details. In this section, we introduce details about self-supervised losses $\mathcal{L}_{\text{self}}$ and knowledge distillation $\mathcal{L}_{\text{kd}}$.

\subsection{Details of Self-Supervised Losses}
Following~\citesupp{feng2021learningsupp,deng2019accuratesupp}, we leverage the differentiable renderer~\citesupp{genova2018unsupervisedsupp} to obtain the rendered image $\hat{\mathbf{I}}^r$, then use photo loss $\mathcal{L}_{\text{pho}}$ and identity loss $\mathcal{L}_{\text{id}}$ to compute the error between the input image $\mathbf{I}$ and $\hat{\mathbf{I}}^r$. We also follow~\citesupp{wood2022densesupp} to use dense landmark loss $\mathcal{L}_{\text{lmk}}$ to calculate the error between the detected landmarks from $\mathbf{I}$ and projected landmarks from $\hat{\mathbf{S}}$, as self-supervised loss is still crucial to ensure satisfactory generalization to real-world images.

Photo loss $\mathcal{L}_{\text{pho}}$ computes the $l_2$ error between $\mathbf{I}$ and $\hat{\mathbf{I}}^{r}$:
\begin{equation}
    \mathcal{L}_{\text{pho}}=\left \|\mathbf{M}_{\mathbf{I}} \odot \left(\mathbf{I} - \hat{\mathbf{I}}^r \right)\right \|_2,
    \label{Eq.photoloss}
\end{equation}
where $\mathbf{M}_{\mathbf{I}}$ is the region-of-interest mask~\citesupp{Zheng2022DecoupledMLsupp} of image $\mathbf{I}$, which only considers facial skins and removes occlusions.

Identity loss $\mathcal{L}_{\text{id}}$ leverages the pretrained face recognition network $\mathbf{\Gamma}$~\citesupp{deng2019arcfacesupp} to estimate the cosine similarity between high-level features from $\mathbf{I}$ and $\hat{\mathbf{I}}^r$:
\begin{equation}
    \mathcal{L}_{\text{id}}=\frac{\mathbf{\Gamma} (\mathbf{I})\cdot \mathbf{\Gamma} (\hat{\mathbf{I}}^r)}{\left \| \mathbf{\Gamma} (\mathbf{I}) \right \|_2\cdot \| \mathbf{\Gamma} (\hat{\mathbf{I}}^r) \|_2},
    \label{Eq.idloss}
\end{equation}

Dense landmark loss $\mathcal{L}_{\text{lmk}}$ leverages the landmark detector~\citesupp{wood2022densesupp} to detect $669$ dense landmarks of given 2D images, and estimate the distance between the detected and projected 2D points from the reconstructed shape $\hat{\mathbf{S}}$:
\begin{equation}
    \mathcal{L}_{\text{lmk}}=\sum_{i=1}^{669} \frac{\left \| \boldsymbol{\mu}_{i} - \hat{\boldsymbol{\mu}}_i\right \|_2 }{2\sigma_{i}^2},
\end{equation}
where $\boldsymbol{\mu}_i$ and $\sigma_i$ are the coordinates and uncertainty of the $i$-th detected landmark from $\mathbf{I}$, respectively. $\hat{\boldsymbol{\mu}}_i$ denotes the $i$-th projected 2D landmark from the reconstructed shape $\hat{\mathbf{S}}$.

\begin{figure}[t!]
    \centering
    \vspace{10pt}
    \begin{overpic}[trim=10.5cm 10cm 8.5cm 6cm,clip,width=1\linewidth,grid=false]{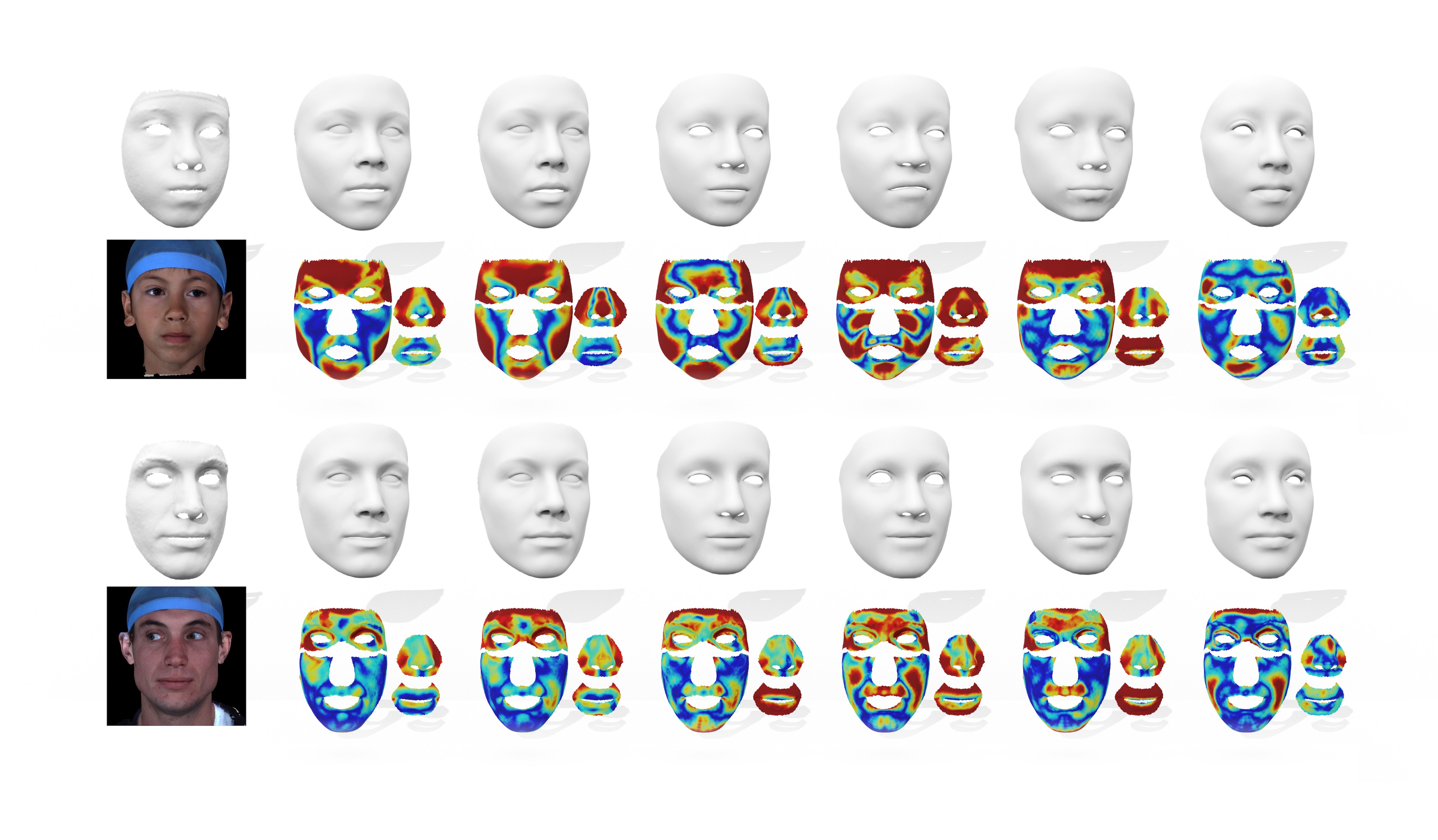}
    \put(3,55){\bfseries\scriptsize G.T.}    
    \put(14,55){\scriptsize Deep3D}
    \put(28,55){\scriptsize MGCNet}
    \put(44,55){\scriptsize DECA}
    \put(58,55){\scriptsize EMOCA}
    \put(74,55){\scriptsize MICA}
    \put(89,55){\bfseries\scriptsize Ours}
    \end{overpic}
    \begin{overpic}[trim=10.5cm 10cm 8.5cm 4cm,clip,width=1\linewidth,grid=false]{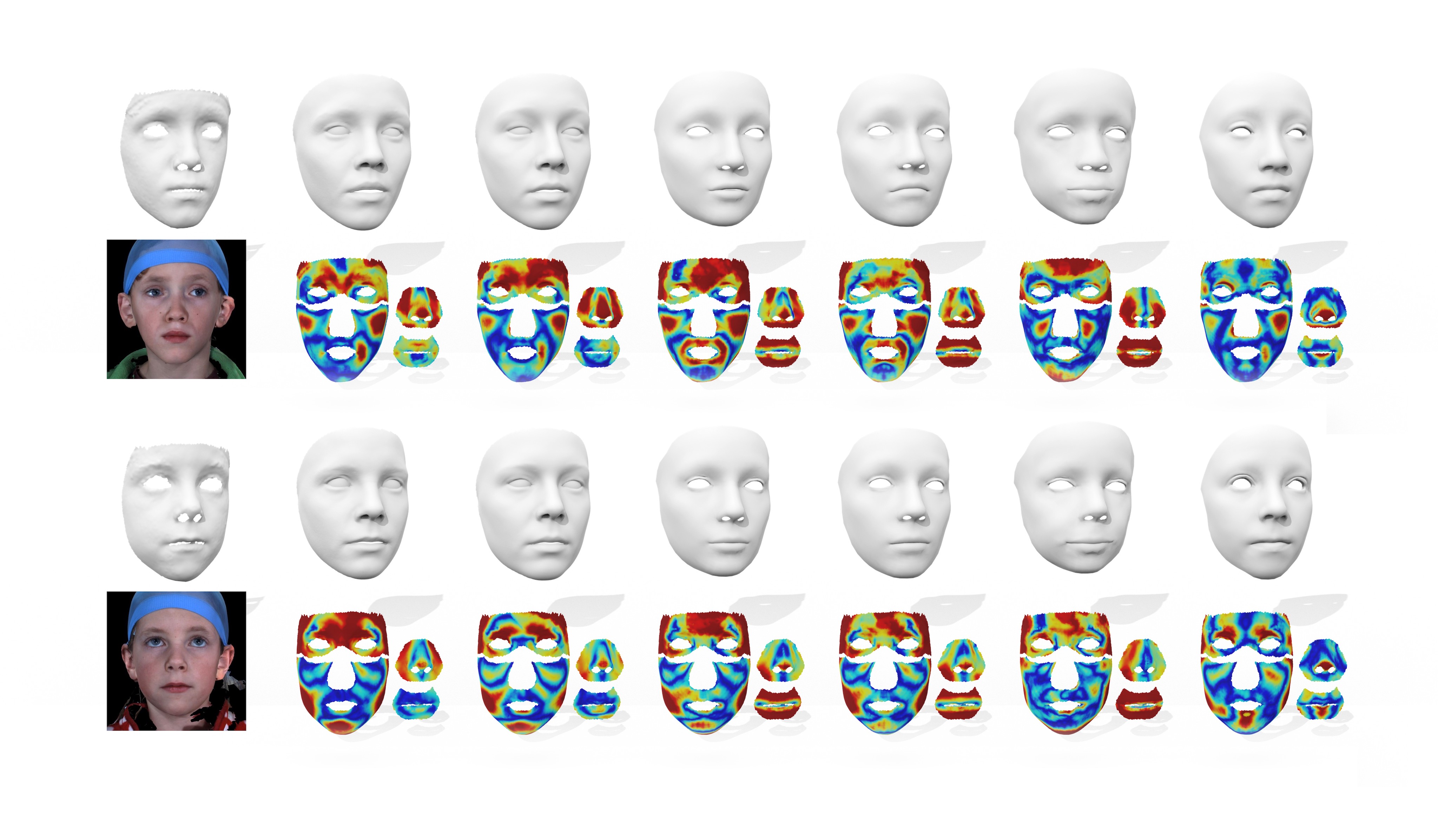}
    \end{overpic}
    \begin{overpic}[trim=10.5cm 10cm 8.5cm 4cm,clip,width=1\linewidth,grid=false]{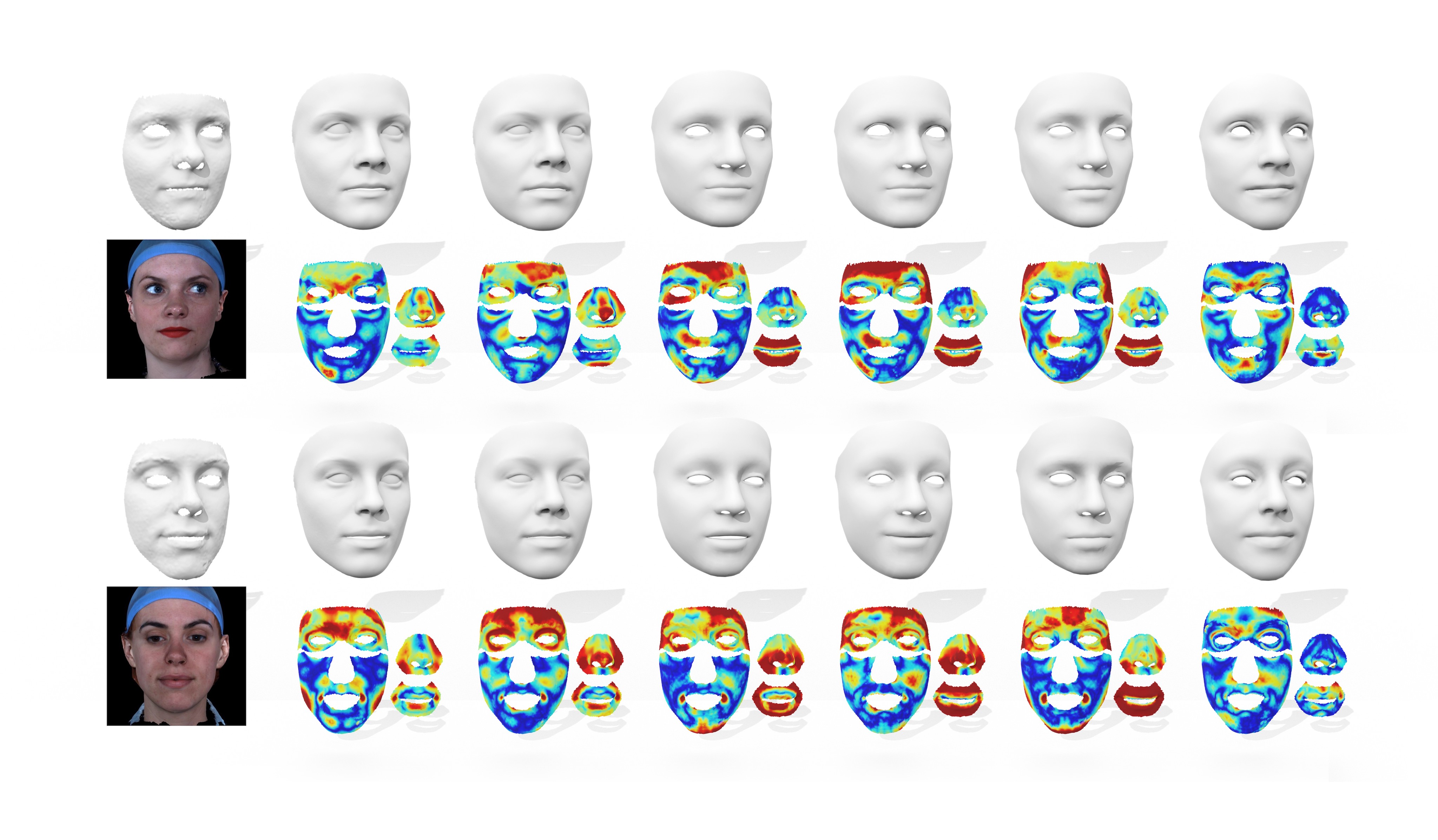}
    \end{overpic}
    \vspace{-5pt}
    \caption{\textbf{Error map on REALY benchmark.} We visualize and compare the reconstruction error of {\name} to previous methods. From left to right: Input image \& ground-truth,  Deep3D~\protect\citesupp{deng2019accuratesupp}, MGCNet~\protect\citesupp{shang2020selfsupp}, DECA~\protect\citesupp{feng2021learningsupp}, EMOCA~\protect\citesupp{danvevcek2022emocasupp}, MICA~\protect\citesupp{MICAsupp}), and {\name} (Ours), where large (small) errors are colored in red (blue). The proposed method presents the best reconstruction quality. } %
    \label{fig:supp_errormap}
    \vspace{-5pt}
\end{figure}

\begin{figure*}[!t]
    \centering
    \begin{overpic}[trim=0cm 0cm 0cm 0cm,clip,width=1\linewidth,grid=false]{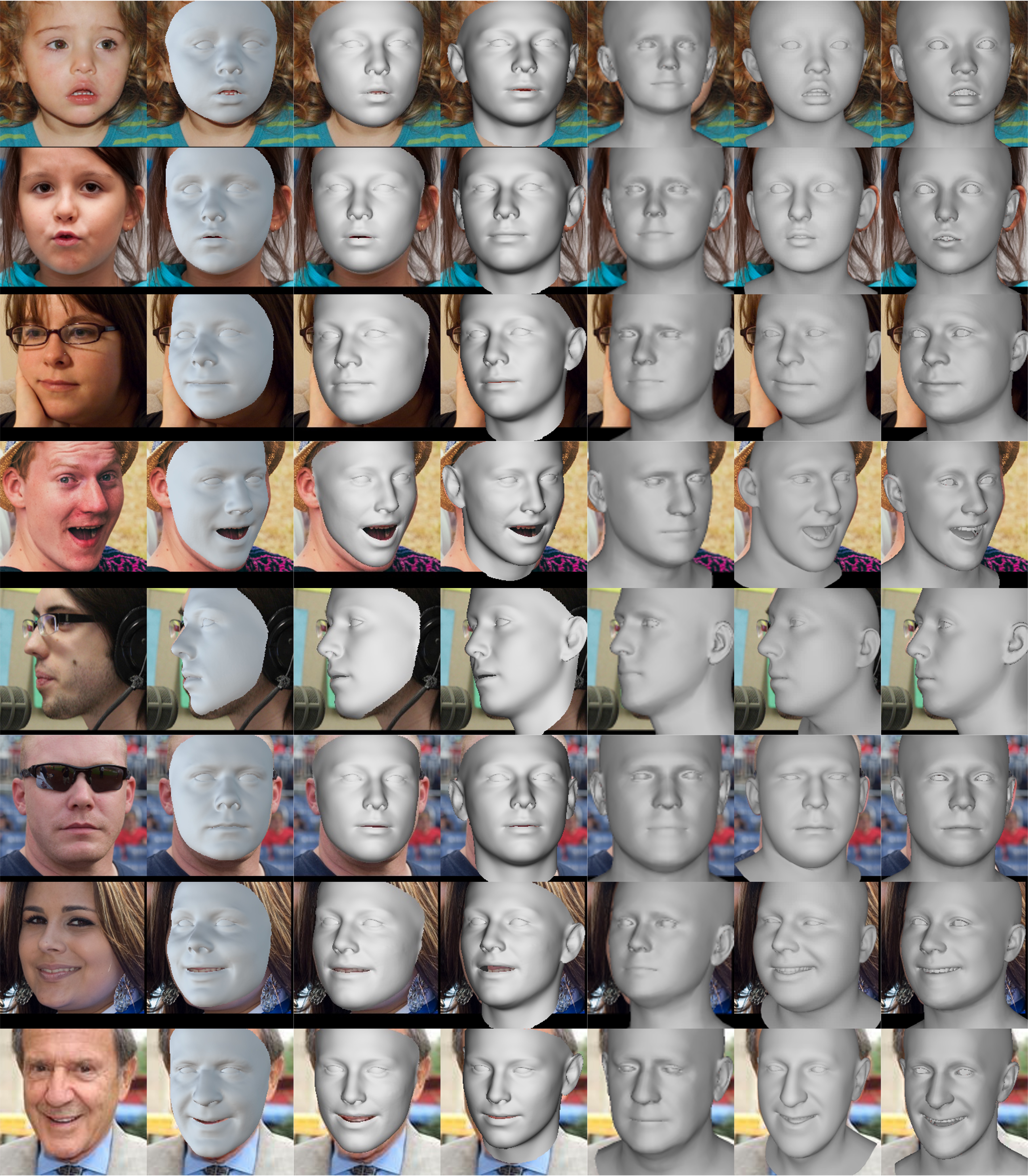}
    \end{overpic}
    \put(-470,570){\bfseries\scriptsize Input}
    \put(-40,570){\bfseries\scriptsize Ours}
    \put(-400,570){\scriptsize Deep3D}
    \put(-330,570){\scriptsize 3DDFA-v2}
    \put(-260,570){\scriptsize SynergyNet} 
    \put(-185,570){\scriptsize MICA} 
    \put(-110,570){\scriptsize Dense}
    \vspace{5pt}
    \caption{\textbf{Comparison on coarse shape reconstruction.} From left to right: Input image, Deep3D~\protect\citesupp{deng2019accuratesupp}, 3DDFA-v2~\protect\citesupp{guo2020towardssupp}, SynergyNet~\protect\citesupp{wu2021synergysupp}, MICA~\protect\citesupp{MICAsupp}, Dense~\protect\citesupp{wood2022densesupp}, and {\name} (Ours). Note that MICA focuses on identity reconstruction, lacking the consideration of expression.}
    \label{fig:supp_coarse_cmp}
    \vspace{-5pt}
\end{figure*}

\begin{figure*}[!t]
    \centering
    \begin{overpic}[trim=0cm 0cm 0cm 0cm,clip,width=1\linewidth,grid=false]{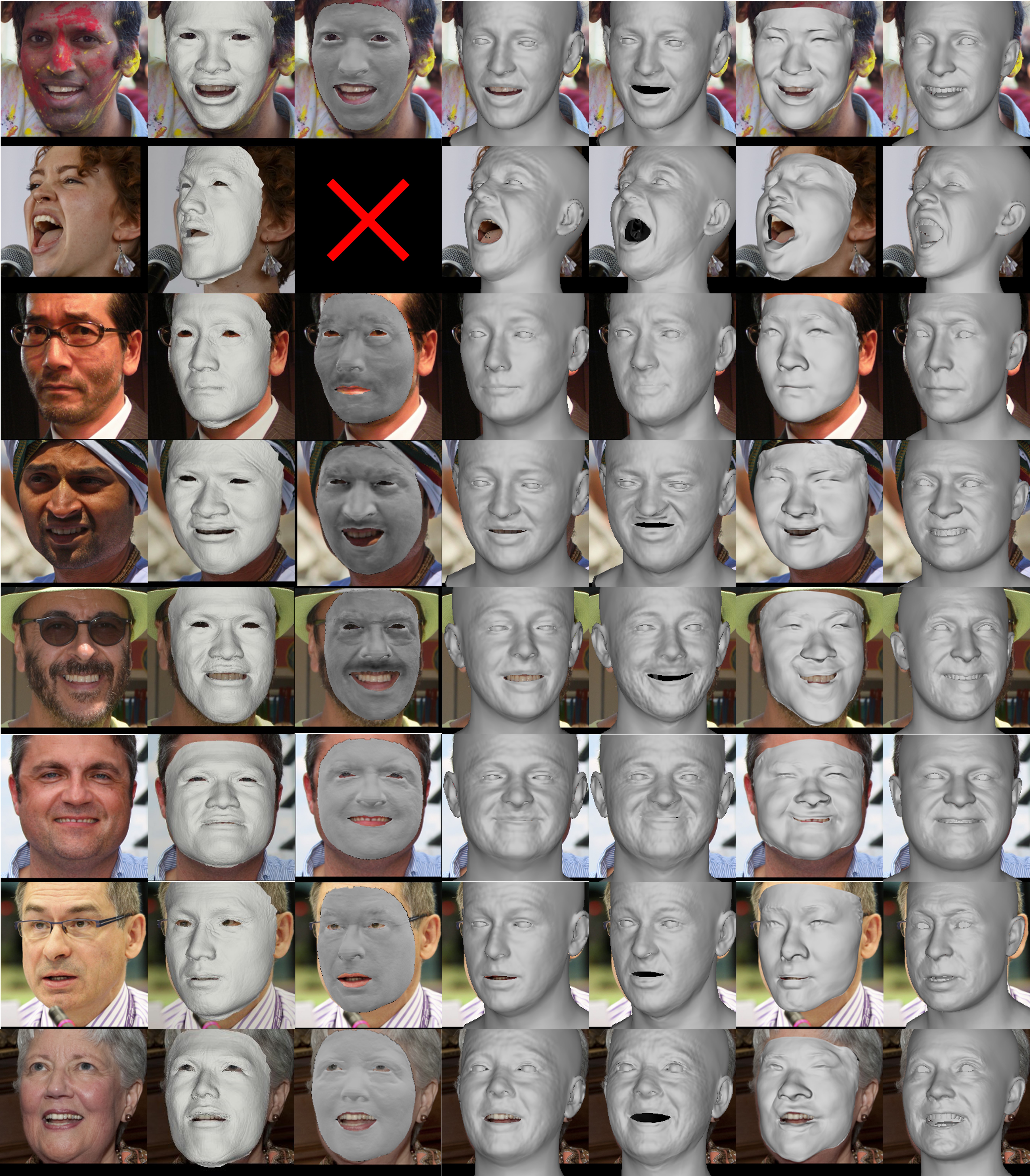}
    \end{overpic}
    \put(-470,570){\bfseries\scriptsize Input}
    \put(-40,570){\bfseries\scriptsize Ours}
    \put(-400,570){\scriptsize FaceScape}
    \put(-327,570){\scriptsize Unsup}
    \put(-255,570){\scriptsize DECA} 
    \put(-185,570){\scriptsize EMOCA} 
    \put(-115,570){\scriptsize FaceVerse}
    \vspace{5pt}
    \caption{\textbf{Comparison on detail shape reconstruction.}  From left to right: Input image, FaceScape~\protect\citesupp{yang2020facescapesupp}, Unsup~\protect\citesupp{chen2020selfsupp}, DECA~\protect\citesupp{feng2021learningsupp}, EMOCA~\protect\citesupp{danvevcek2022emocasupp}, FaceVerse~\protect\citesupp{wang2022faceversesupp}, and {\name} (Ours). ``{\XBox}'' indicates this method fails to return any reconstruction.}
    \label{fig:supp_detail_cmp}
    \vspace{-5pt}
\end{figure*}

\begin{figure*}[ht!]
    \centering
    \begin{overpic}[trim=0cm 0cm 0cm 0cm,clip,width=1\linewidth,grid=false]{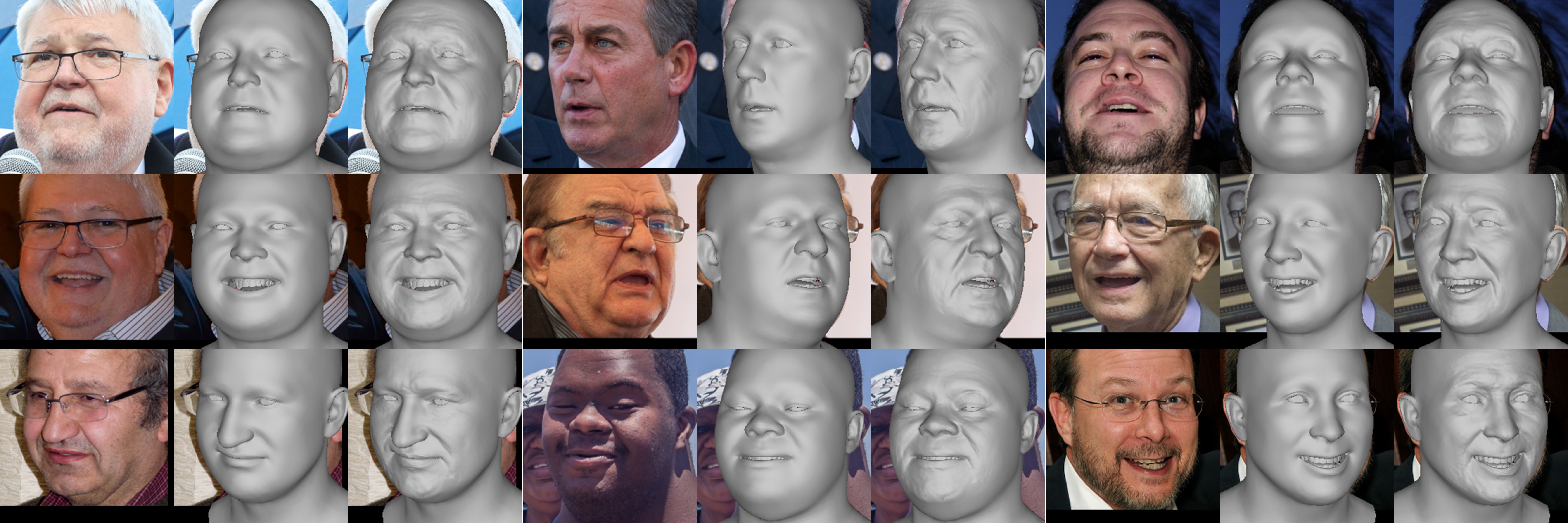}
    \end{overpic}
    \put(-480,168){\bfseries\scriptsize Input}
    \put(-430,168){\scriptsize Dense~\citesupp{wood2022densesupp}}  
    \put(-380,168){\bfseries\scriptsize + Our Detail}   
    \put(-310,168){\bfseries\scriptsize Input}
    \put(-260,168){\scriptsize Dense~\citesupp{wood2022densesupp}}  
    \put(-210,168){\bfseries\scriptsize + Our Detail}   
    \put(-145,168){\bfseries\scriptsize Input}
    \put(-95,168){\scriptsize  Dense~\citesupp{wood2022densesupp}}  
    \put(-45,168){\bfseries\scriptsize + Our Detail}  
    \vspace{-5pt}
    \caption{\textbf{Illustration on the flexibility of {\module}.} Given the identity and expression coefficients ($\boldsymbol{\beta}$, $\boldsymbol{\xi}$) from the optimization-based method~\protect\citesupp{wood2022densesupp}, {\module} can generate realistic details based on the coarse shape and further improve the visual quality.}
    \label{fig:supp_ft_detail}
\end{figure*}

\subsection{Details of Knowledge Distillation}
The age prediction model $\mathbf{\Gamma}_{\text{age}}$~\citesupp{karkkainenfairfacesupp} predicts the age of given images into $9$ categories: $0-2$, $3-9$, $10-19$, $20-29$, $30-39$, $40-49$, $50-59$, $60-69$, and $70+$. Therefore, we leverage $3$-layer MLP to transform the static coefficient $\boldsymbol{\varphi}$ into a $9$-dim vector, and use \textit{softmax} to map into a probability distribution $\hat{\mathbf{p}}_{\text{age}}$.

\section{Additional Experimental Results} \label{sec.supp_exp}

In this section, we provide additional results to strengthen the superiority of {\name} in reconstructing 3D shapes with animatable details. Specifically, we present 1). error maps on REALY~\citesupp{REALYsupp} in Fig.~\ref{fig:supp_errormap}, 2). additional reconstruction comparisons of coarse shape and details in Fig.~\ref{fig:supp_coarse_cmp} and Fig.~\ref{fig:supp_detail_cmp}, respectively, 3). additional experiments on flexibility for both images and video sequences in Fig.~\ref{fig:supp_ft_detail} and Fig.~\ref{fig:supp_ft_detail_video1}, Fig.~\ref{fig:supp_ft_detail_video2},  Fig.~\ref{fig:supp_ft_detail_video3}, respectively, and 4). additional animation comparisons in Fig.~\ref{fig:supp_animation} and Fig.~\ref{fig:supp_animation2}, 5). user studies about the reconstruction and animation quality in Tab.~\ref{tab:userstudy} and Tab.~\ref{tab:supp_user2}, respectively.

\subsection{Error Maps of REALY Benchmark}

In Fig.~\ref{fig:supp_errormap}, we present additional comparisons of the error map on the REALY benchmark~\citesupp{REALYsupp}. The RGB color in ground-truth regions is mapped from the vertex-to-vertex error between the ground-truth and predicted shape according to the evaluation protocol in~\citesupp{REALYsupp}. Compared to previous methods, {\name} reaches the smallest error and the best reconstruction quality.

\subsection{Reconstruction Comparisons}

\myparagraph{Qualitative Comparisons.}
In Fig.~\ref{fig:supp_coarse_cmp} and Fig.~\ref{fig:supp_detail_cmp}, we present additional comparisons of {\name} to previous coarse shape reconstruction methods~\citesupp{deng2019accuratesupp,guo2020towardssupp,wu2021synergysupp,MICAsupp,wood2022densesupp} and detail reconstruction methods~\citesupp{yang2020facescapesupp,chen2020selfsupp,feng2021learningsupp,danvevcek2022emocasupp,wang2022faceversesupp}. The input images show diversity {\wrt} ethnicity, gender, age, BMI, pose, environment, occlusion, and expression.
Compared to previous methods, {\name} is robust to occlusions, extreme poses, and diversity expressions. {\name} reconstructs realistic coarse shapes, better expressions, and realistic details.

\begin{table}[ht!]
\caption{\textbf{User study results on the reconstructed shape and details.} {\name} achieves the best results in coarse shape and details according to human perception compared to prior art~\protect\citesupp{feng2021learningsupp,wood2022densesupp}.}\label{tab:userstudy}
\vspace{5pt}
\resizebox{1\linewidth}{!}{%
\begin{tabular}{c|ccc}
\toprule[1pt]
\cellcolor{tabblue!10} Group & \cellcolor{red!10}{Best method}  & \cellcolor{tabblue!10} 2nd best method &\cellcolor{red!10} Other methods \\ \midrule[1pt]
Coarse         & \textbf{54.55\%} (Ours) & 35.06\% (Dense~\citesupp{wood2022densesupp})   & 10.39\% (\citesupp{MICAsupp,deng2019accuratesupp,wu2021synergysupp,guo2020towardssupp})      \\ \midrule[1pt]
Detail         & \textbf{80.52\%} (Ours) & 11.69\% (DECA~\citesupp{feng2021learningsupp})    & 7.79\% (\citesupp{danvevcek2022emocasupp,wang2022faceversesupp,chen2020selfsupp,yang2020facescapesupp})       \\ \bottomrule[1pt]
\end{tabular}
}
\vspace{-5pt}
\end{table}

\myparagraph{User Study.}
To demonstrate that our reconstructed 3D faces present visually better results and are faithfully aligned with human perception, we present a user study by inviting $77$ volunteers with a computer science background to vote for the best-reconstructed shapes, from sampled faces in CelebA~\citesupp{CelebAMask-HQsupp}, FFHQ~\citesupp{karras2019stylesupp}, and AFLW2000~\citesupp{yin2017towardssupp}. Specifically, we separately compare methods for coarse shape reconstruction~\citesupp{deng2019accuratesupp,guo2020towardssupp,wu2021synergysupp,MICAsupp,wood2022densesupp}, and detailed reconstruction~\citesupp{yang2020facescapesupp,chen2020selfsupp,feng2021learningsupp,danvevcek2022emocasupp,wang2022faceversesupp}. The results are summarized in Tab.~\ref{tab:userstudy}.

\begin{figure*}[t!]
    \centering
    \begin{overpic}[trim=0cm 0cm 0cm 0cm,clip,width=1\linewidth,grid=false]{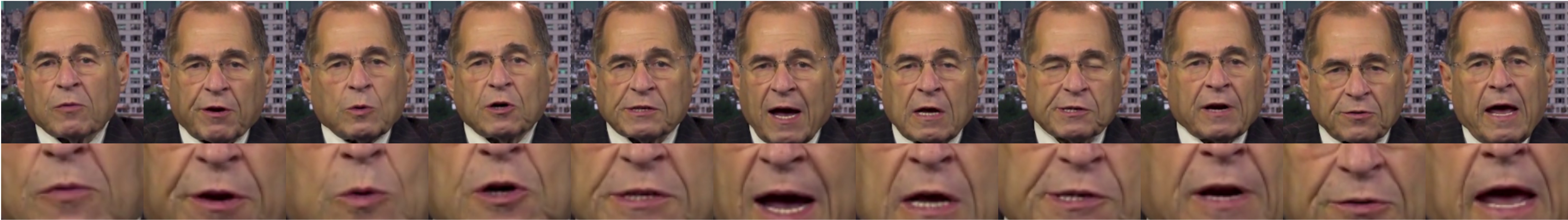}
    \end{overpic}
    \put(-505,40){\bfseries\scriptsize \rotatebox{90}{Input}}
    \put(-480,71){\bfseries\scriptsize T=0}
    \put(-30,71){\bfseries\scriptsize T=10}
    \put(-75,71){\bfseries\scriptsize T=9}
    \put(-120,71){\bfseries\scriptsize T=8}
    \put(-165,71){\bfseries\scriptsize T=7}
    \put(-210,71){\bfseries\scriptsize T=6}
    \put(-255,71){\bfseries\scriptsize T=5}
    \put(-300,71){\bfseries\scriptsize T=4}
    \put(-345,71){\bfseries\scriptsize T=3}
    \put(-390,71){\bfseries\scriptsize T=2}
    \put(-435,71){\bfseries\scriptsize T=1}
    
    \begin{overpic}[trim=0cm 0cm 0cm 0cm,clip,width=1\linewidth,grid=false]{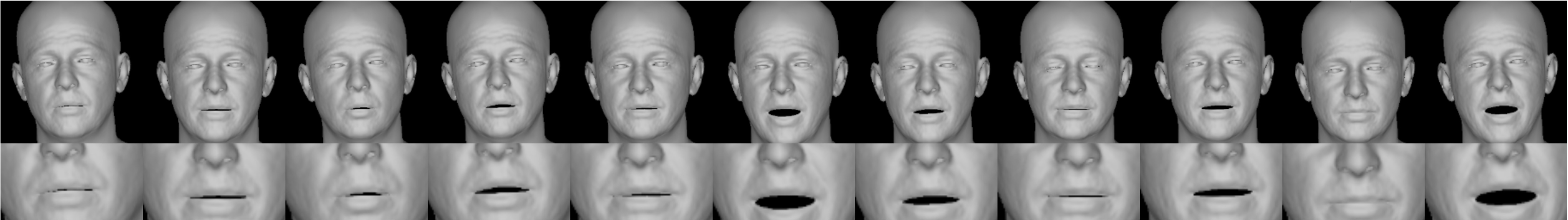}
    \end{overpic}
    \put(-505,35){\bfseries\scriptsize \rotatebox{90}{DECA~\citesupp{feng2021learningsupp}}}

    \begin{overpic}[trim=0cm 0cm 0cm 0cm,clip,width=1\linewidth,grid=false]{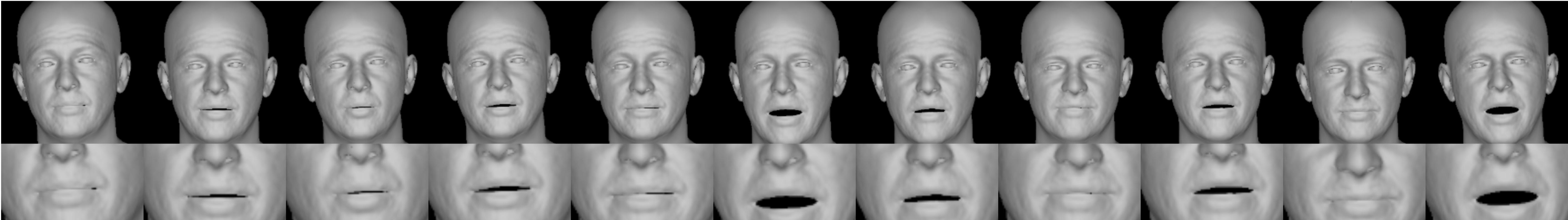}
    \end{overpic}
    \put(-505,30){\bfseries\scriptsize \rotatebox{90}{EMOCA~\citesupp{danvevcek2022emocasupp}}}

    \begin{overpic}[trim=0cm 0cm 0cm 0cm,clip,width=1\linewidth,grid=false]{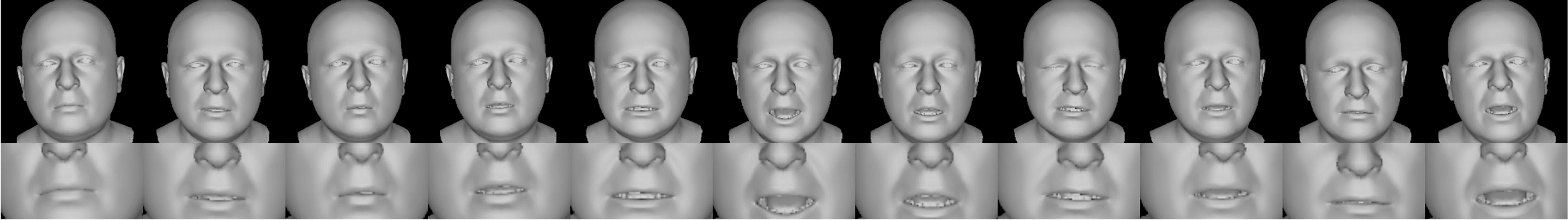}
    \end{overpic}
    \put(-505,35){\bfseries\scriptsize \rotatebox{90}{Dense~\citesupp{wood2022densesupp}}}
    
    \begin{overpic}[trim=0cm 0cm 0cm 0cm,clip,width=1\linewidth,grid=false]{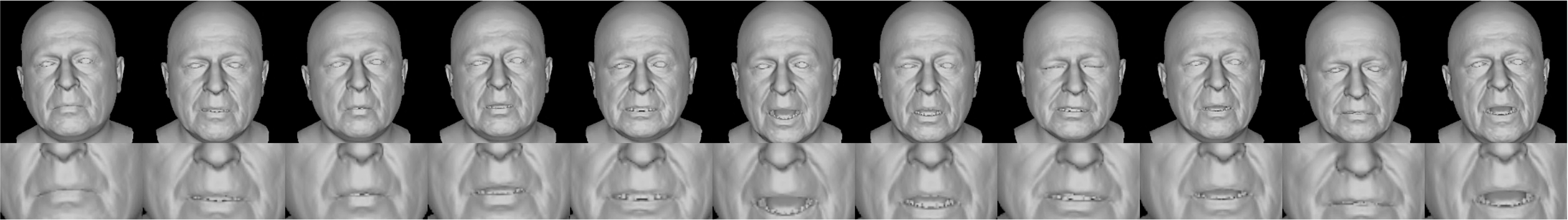}
    \end{overpic}
    \put(-505,40){\bfseries\scriptsize \rotatebox{90}{Ours}}
    \vspace{-5pt}
    \caption{\textbf{Illustration on the flexibility of {\module} on video reconstruction (part 1).} We visualized the reconstruction quality of Dense~\protect\citesupp{wood2022densesupp} with/without our {\module} and compare them with prior art~\protect\citesupp{feng2021learningsupp,danvevcek2022emocasupp}. 
    Videos are taken from YouTube.
    }
    \label{fig:supp_ft_detail_video1}
    \vspace{-8pt}
\end{figure*}

\begin{figure*}[t!]
    \centering
    \begin{overpic}[trim=0cm 0cm 0cm 0cm,clip,width=1\linewidth,grid=false]{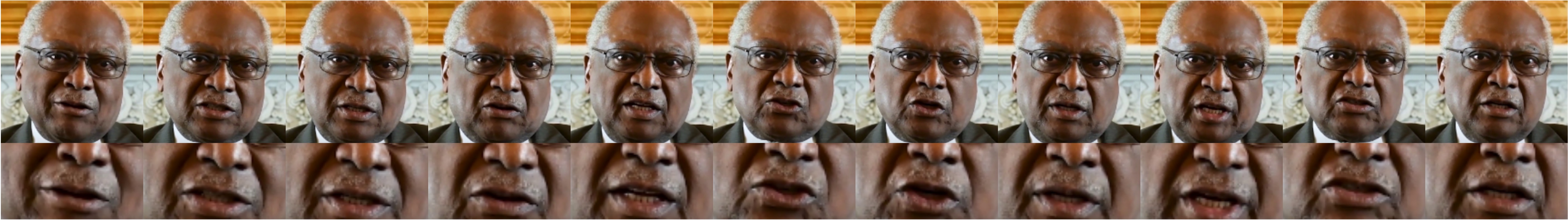}
    \end{overpic}
    \put(-505,40){\bfseries\scriptsize \rotatebox{90}{Input}}
    \put(-480,71){\bfseries\scriptsize T=0}
    \put(-30,71){\bfseries\scriptsize T=10}
    \put(-75,71){\bfseries\scriptsize T=9}
    \put(-120,71){\bfseries\scriptsize T=8}
    \put(-165,71){\bfseries\scriptsize T=7}
    \put(-210,71){\bfseries\scriptsize T=6}
    \put(-255,71){\bfseries\scriptsize T=5}
    \put(-300,71){\bfseries\scriptsize T=4}
    \put(-345,71){\bfseries\scriptsize T=3}
    \put(-390,71){\bfseries\scriptsize T=2}
    \put(-435,71){\bfseries\scriptsize T=1}
    
    \begin{overpic}[trim=0cm 0cm 0cm 0cm,clip,width=1\linewidth,grid=false]{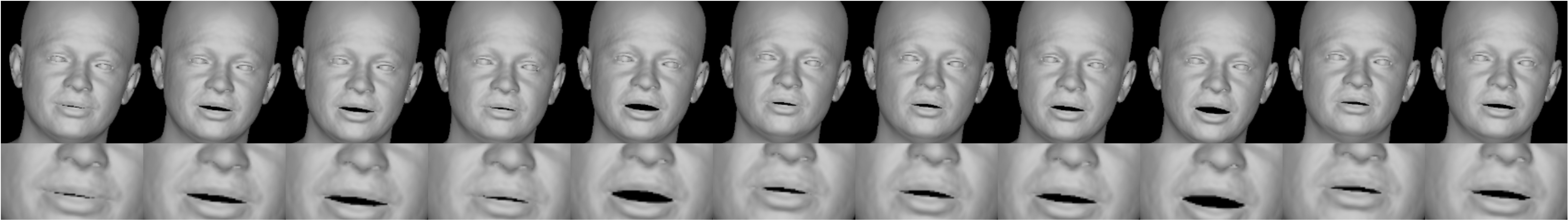}
    \end{overpic}
    \put(-505,35){\bfseries\scriptsize \rotatebox{90}{DECA~\citesupp{feng2021learningsupp}}}

    \begin{overpic}[trim=0cm 0cm 0cm 0cm,clip,width=1\linewidth,grid=false]{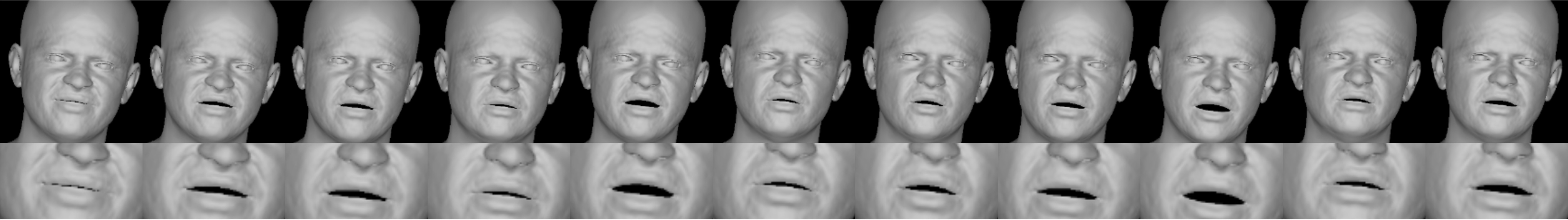}
    \end{overpic}
    \put(-505,30){\bfseries\scriptsize \rotatebox{90}{EMOCA~\citesupp{danvevcek2022emocasupp}}}

    \begin{overpic}[trim=0cm 0cm 0cm 0cm,clip,width=1\linewidth,grid=false]{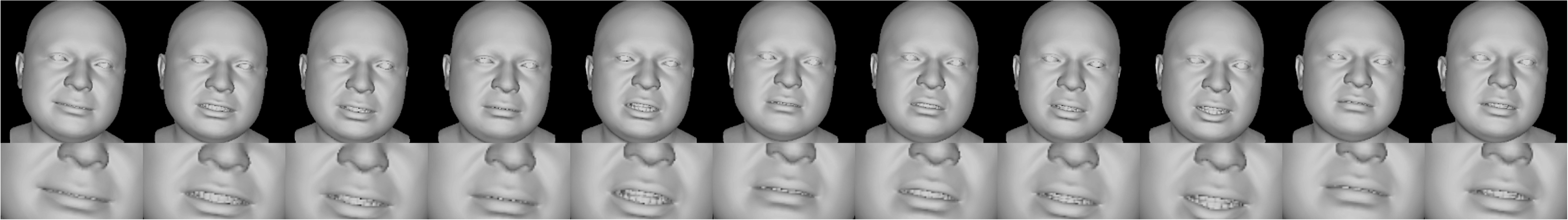}
    \end{overpic}
    \put(-505,35){\bfseries\scriptsize \rotatebox{90}{Dense~\citesupp{wood2022densesupp}}}
    
    \begin{overpic}[trim=0cm 0cm 0cm 0cm,clip,width=1\linewidth,grid=false]{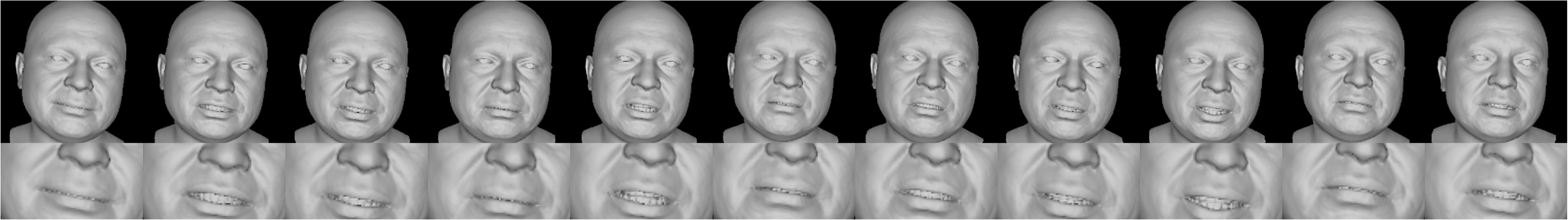}
    \end{overpic}
    \put(-505,40){\bfseries\scriptsize \rotatebox{90}{Ours}}
    \vspace{-5pt}
    \caption{\textbf{Illustration on the flexibility of {\module} on video reconstruction (part 2).} We visualized the reconstruction quality of Dense~\protect\citesupp{wood2022densesupp} with/without our {\module} and compare them with prior art~\protect\citesupp{feng2021learningsupp,danvevcek2022emocasupp}. 
    Videos are taken from YouTube.
    }
    \label{fig:supp_ft_detail_video2}
\end{figure*}

\begin{figure*}[t!]
    \centering
    \begin{overpic}[trim=0cm 0cm 0cm 0cm,clip,width=1\linewidth,grid=false]{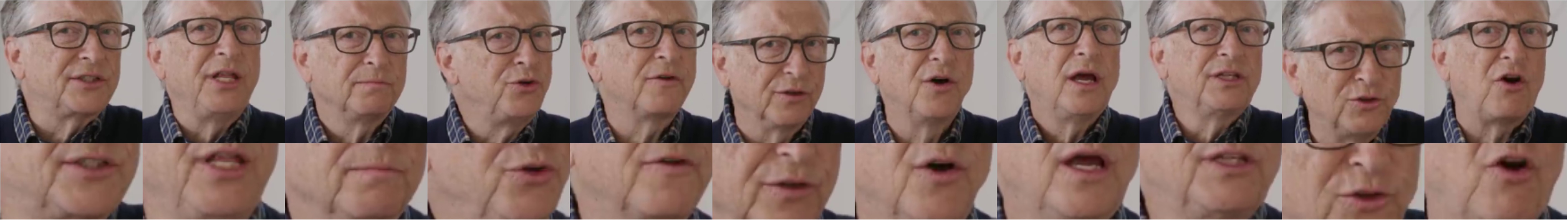}
    \end{overpic}
    \put(-505,40){\bfseries\scriptsize \rotatebox{90}{Input}}
    \put(-480,71){\bfseries\scriptsize T=0}
    \put(-30,71){\bfseries\scriptsize T=10}
    \put(-75,71){\bfseries\scriptsize T=9}
    \put(-120,71){\bfseries\scriptsize T=8}
    \put(-165,71){\bfseries\scriptsize T=7}
    \put(-210,71){\bfseries\scriptsize T=6}
    \put(-255,71){\bfseries\scriptsize T=5}
    \put(-300,71){\bfseries\scriptsize T=4}
    \put(-345,71){\bfseries\scriptsize T=3}
    \put(-390,71){\bfseries\scriptsize T=2}
    \put(-435,71){\bfseries\scriptsize T=1}
    
    \begin{overpic}[trim=0cm 0cm 0cm 0cm,clip,width=1\linewidth,grid=false]{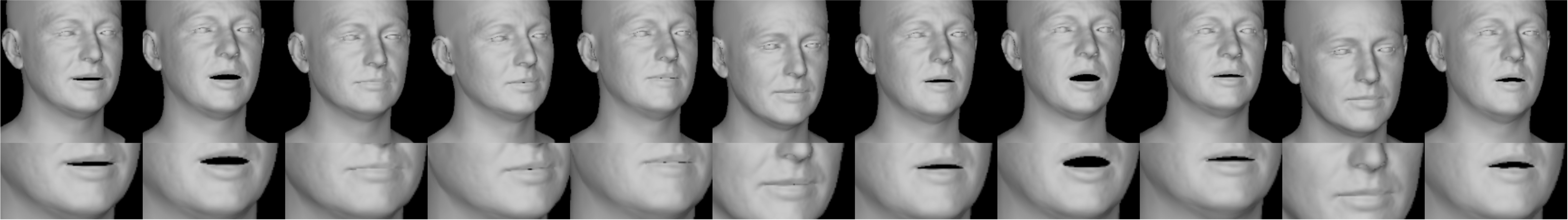}
    \end{overpic}
    \put(-505,35){\bfseries\scriptsize \rotatebox{90}{DECA~\citesupp{feng2021learningsupp}}}

    \begin{overpic}[trim=0cm 0cm 0cm 0cm,clip,width=1\linewidth,grid=false]{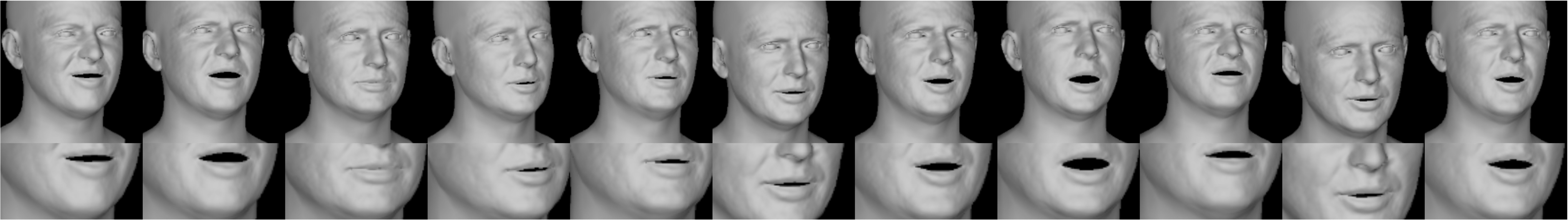}
    \end{overpic}
    \put(-505,30){\bfseries\scriptsize \rotatebox{90}{EMOCA~\citesupp{danvevcek2022emocasupp}}}

    \begin{overpic}[trim=0cm 0cm 0cm 0cm,clip,width=1\linewidth,grid=false]{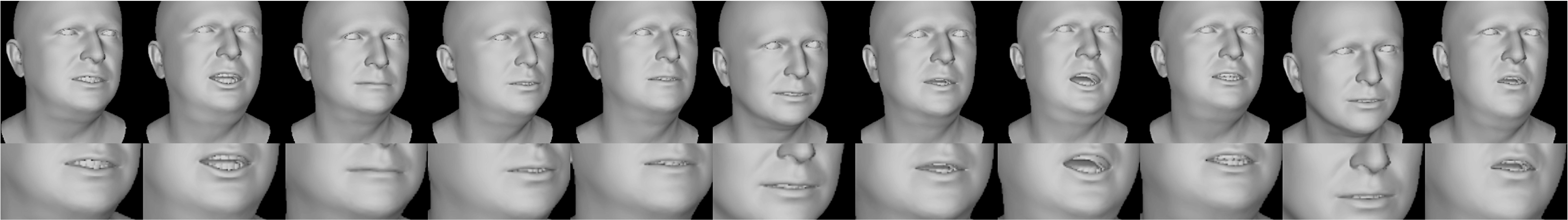}
    \end{overpic}
    \put(-505,35){\bfseries\scriptsize \rotatebox{90}{Dense~\citesupp{wood2022densesupp}}}
    
    \begin{overpic}[trim=0cm 0cm 0cm 0cm,clip,width=1\linewidth,grid=false]{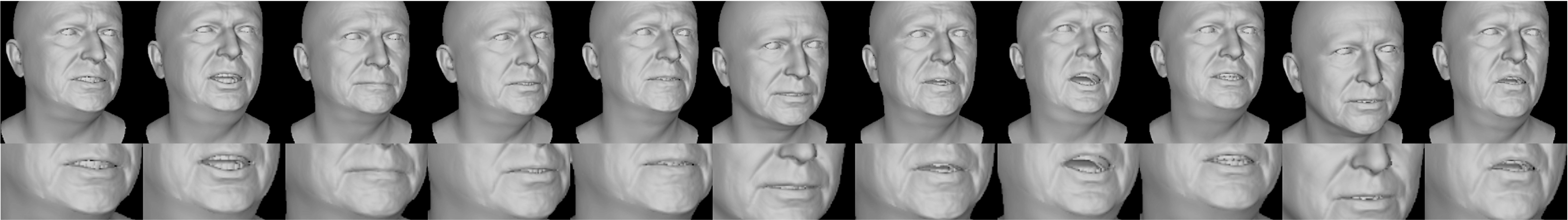}
    \end{overpic}
    \put(-505,40){\bfseries\scriptsize \rotatebox{90}{Ours}}
    \vspace{-5pt}
    \caption{\textbf{Illustration on the flexibility of {\module} on video reconstruction (part 3).} We visualized the reconstruction quality of Dense~\protect\citesupp{wood2022densesupp} with/without our {\module} and compare them with prior art~\protect\citesupp{feng2021learningsupp,danvevcek2022emocasupp}.
    Videos are taken from YouTube.
    }
    \label{fig:supp_ft_detail_video3}
\end{figure*}

Tab.~\ref{tab:userstudy} shows that more than half of users perceive that our reconstructed shapes are more similar to the given images, and $80.52\%$ users vote that {\name} reconstructs more realistic details compared to others. As a comparison, the second-best detail reconstruction method~\citesupp{feng2021learningsupp} has only $11.69\%$ votes. It demonstrates the superiority of our methods in reconstructing coarse shapes and details.

\begin{table*}[t!]
\caption{\textbf{User study results on detail and expression transfer.} We group the driving images into young people (A) and elder people (B), and ask artists to score the animation quality of static details and dynamic expressions of the transferred images, corresponding to the 1st and 2nd row of each source image in Fig.~\ref{fig:supp_animation} and Fig.~\ref{fig:supp_animation2}. We report the average scores, median scores, and standard deviation. ``5'' indicates the best score while ``1'' indicates the worst score.
}
\label{tab:supp_user2}
\vspace{5pt}
\resizebox{1\linewidth}{!}{%
\begin{tabular}{cc|cccccc|cccccc|cccccc}
\toprule[1pt]
\multicolumn{2}{c|}{\cellcolor{red!10}{Driving Group}}                & \multicolumn{6}{c|}{\cellcolor{tabblue!10}{A (1-5)}}  & \multicolumn{6}{c|}{\cellcolor{red!10}{B (6-10)}}  & \multicolumn{6}{{c}}{\cellcolor{tabblue!10}{All}} \\ \midrule[1pt]
\multicolumn{1}{c|}{\multirow{2}{*}{Source}}     &  \multirow{2}{*}{Method} & \multicolumn{3}{c}{Static}   & \multicolumn{3}{c|}{Dynamic}   & \multicolumn{3}{c}{Static}   & \multicolumn{3}{c|}{Dynamic}   & \multicolumn{3}{c}{Static}   & \multicolumn{3}{c}{Dynamic}   \\ \cline{3-20}
\multicolumn{1}{c|}{}                                &                         & avg.   & med.   & std.   & avg.    & med.   & std.   & avg.   & med.   & std.   & avg.  & med.  & std.  & avg.   & med.   & std.   & avg.   & med.   & std.   \\ \midrule[1pt]
\multicolumn{1}{c|}{\multirow{3}{*}{I}} & DECA~\citesupp{feng2021learningsupp}   &  1.63 & 1.50 & 0.72      &  2.50 & 2.50 & 1.05      & 3.27 & 3.00 & 0.65     &   2.73 & 2.50 & 0.80     &          2.45 & 2.50 & 1.07      &  2.62 & 2.50 & 0.93    \\
\multicolumn{1}{c|}{}                   & EMOCA~\citesupp{danvevcek2022emocasupp}  &  1.93 & 2.00 & 0.82      &    1.97 & 2.00 & 0.92     &   3.20 & 3.00 & 0.62   & 2.40 & 2.50 & 0.74       &   2.57 & 3.00 & 0.96      & 2.18 & 2.00 & 0.85       \\
\multicolumn{1}{c|}{}                   & Ours   &   \textbf{4.40} & \textbf{4.50} & 0.60     &   \textbf{4.63} & \textbf{5.00} & 0.44            &   \textbf{4.13} &  \textbf{4.00} & 0.35   &   \textbf{4.53} & \textbf{4.50} & 0.44     &   \textbf{4.27} & \textbf{4.00} & 0.50      &   \textbf{4.58} & \textbf{4.50} & 0.44   \\ \midrule[1pt]
\multicolumn{1}{c|}{\multirow{3}{*}{II}} & DECA~\citesupp{feng2021learningsupp}   &   1.70 & 2.00 & 0.62     &    2.47 & 2.00 & 1.08     & 3.37 & 3.00 & 0.52     &  3.10 & 3.00 & 0.78      &   2.53 & 2.75 & 1.02      & 2.78 & 3.00 & 0.98       \\
\multicolumn{1}{c|}{}                   & EMOCA~\citesupp{danvevcek2022emocasupp}  &   2.07 & 2.00 & 0.73     &   2.37 & 2.00 & 1.03     &  3.73 & 4.00 & 0.68     & 2.07 & 2.00 & 0.84       &   2.90 & 3.00 & 1.09      &  2.22 & 2.00 & 0.90       \\
\multicolumn{1}{c|}{}                   & Ours   &   \textbf{4.17} & \textbf{4.00} & 0.49     &  \textbf{4.80} & \textbf{5.00} & 0.41       &     \textbf{4.30} & \textbf{4.00} & 0.59     &   \textbf{4.47} & \textbf{4.00} & 0.40     &  \textbf{4.23} & \textbf{4.00} & 0.54       &   \textbf{4.63} & \textbf{5.00} & 0.43   \\ \bottomrule[1pt]
\end{tabular}
}
\end{table*}

\begin{figure*}[!t]
    \centering
    \begin{overpic}[trim=0cm 0cm 0cm 0cm,clip,width=1\linewidth,grid=false]{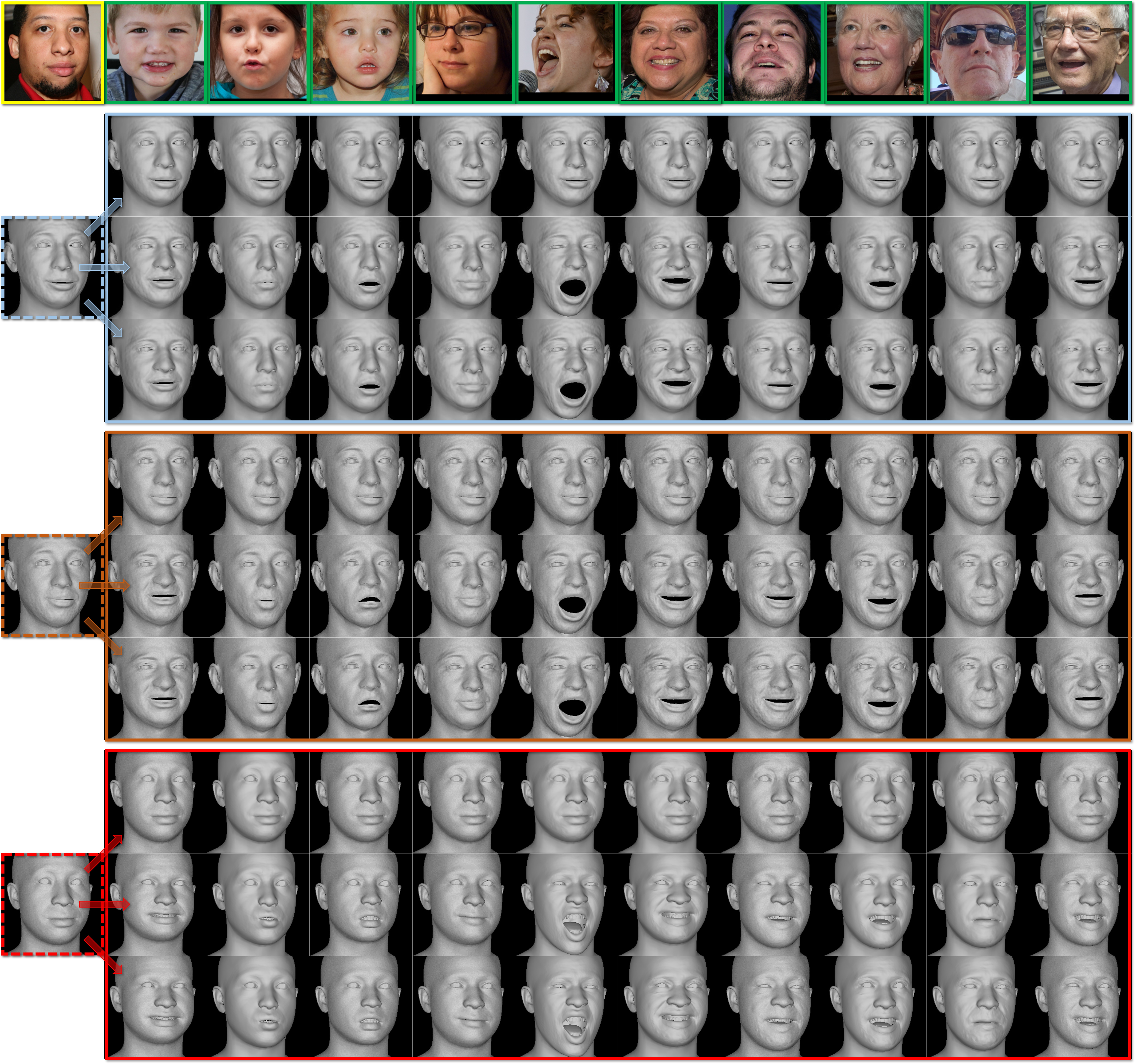}
    \end{overpic}
    \put(-483,38){\bfseries\scriptsize Ours}    
    \put(-493,178){\bfseries\scriptsize EMOCA~\citesupp{danvevcek2022emocasupp}}   
    \put(-490,318){\bfseries\scriptsize DECA~\citesupp{feng2021learningsupp}} 
    \vspace{-5pt}
    \caption{\textbf{Comparison on face animation (part 1).} Given a source image (yellow box), we use the driving images (green box) to drive its person-specific details and expressions. For each method, we manipulate the static (1st-row), dynamic (2nd-row), or both (3rd-row) factors. However, DECA~\protect\citesupp{feng2021learningsupp} (blue box) and EMOCA~\protect\citesupp{danvevcek2022emocasupp} (orange box) can animate the expression-driven details but lack realistic, and cannot transfer the static details from the driving images well. As a comparison, {\name} (red box) is flexible to animate details from static, dynamic, or both factors, and presents vivid animation quality with realistic shapes.
    }
    \label{fig:supp_animation}
\end{figure*}

\begin{figure*}[!t]
    \centering
    \begin{overpic}[trim=0cm 0cm 0cm 0cm,clip,width=1\linewidth,grid=false]{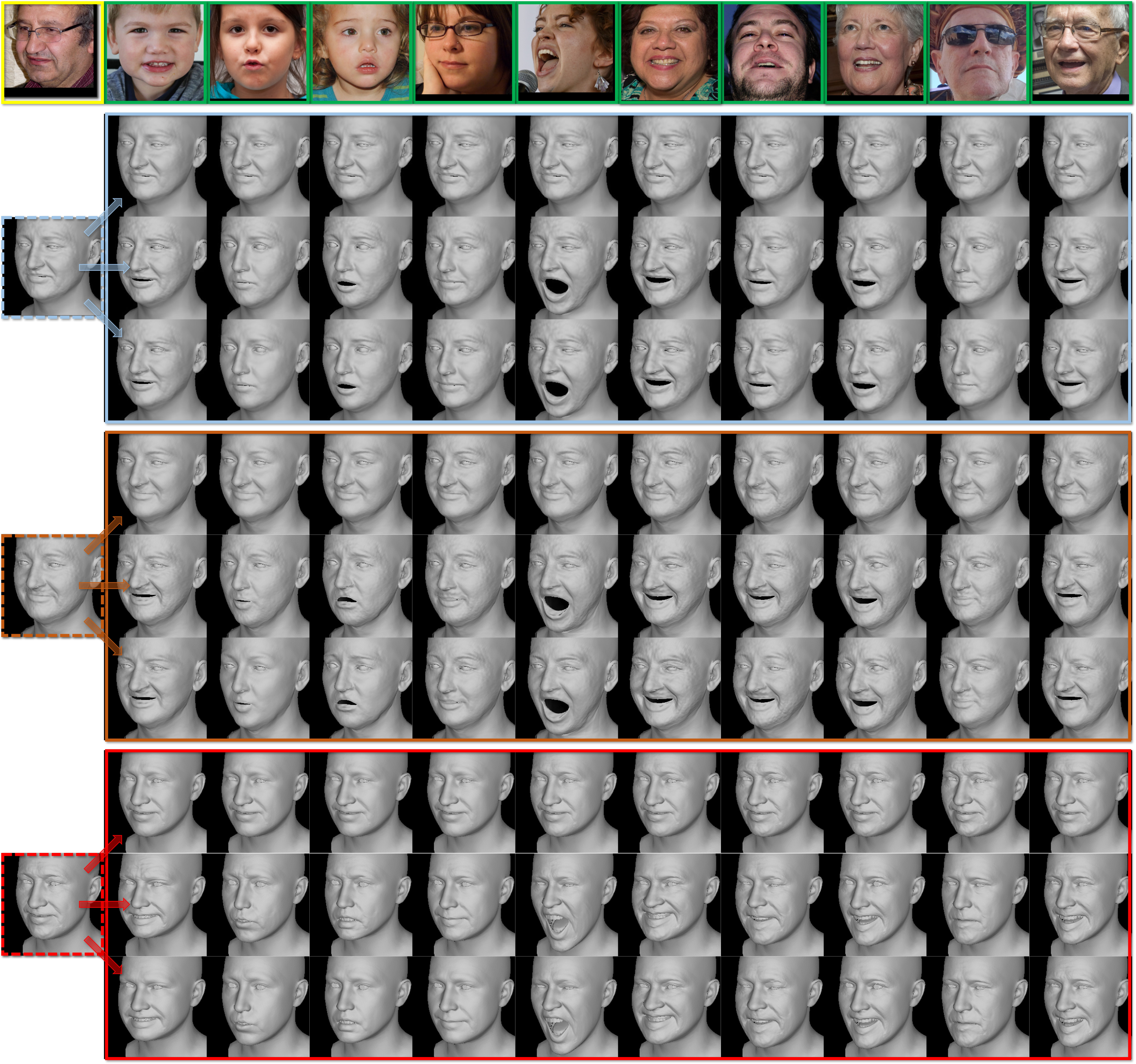}
    \end{overpic}
    \put(-483,38){\bfseries\scriptsize Ours}    
    \put(-493,178){\bfseries\scriptsize EMOCA~\citesupp{danvevcek2022emocasupp}}   
    \put(-490,318){\bfseries\scriptsize DECA~\citesupp{feng2021learningsupp}}  
    \vspace{-5pt}
    \caption{\textbf{Comparison on face animation (part 2).} Given a source image (yellow box), we use the driving images (green box) to drive its person-specific details and expressions. For each method, we manipulate the static (1st-row), dynamic (2nd-row), or both (3rd-row) factors. However, DECA~\protect\citesupp{feng2021learningsupp} (blue box) and EMOCA~\protect\citesupp{danvevcek2022emocasupp} (orange box) can animate the expression-driven details but lack realistic, and cannot transfer the static details from the driving images well. As a comparison, {\name} (red box) is flexible to animate details from static, dynamic, or both factors, and presents vivid animation quality with realistic shapes.
    }
    \vspace{-1pt}
    \label{fig:supp_animation2}
\end{figure*}

\subsection{Flexibility of {\module}}

We introduce the settings of plugging {\module} into optimization-based methods such as Dense~\citesupp{wood2022densesupp}. Specifically, for a given image, we leverage optimization-based methods to regress the identity coefficient $\boldsymbol{\beta}$ and expression coefficient $\boldsymbol{\xi}$, and use our feature extractor in {\name} to regress the static coefficient $\boldsymbol{\varphi}$. Then these three coefficients serve as the input of {\module} to synthesize the realistic displacement maps. Finally, we integrate the coarse shape from the optimization-based methods with the synthesized displacement map to obtain the final detailed shapes. 
In Fig.~\ref{fig:supp_ft_detail}, we present additional results to justify the flexibility of {\module}. {\module} can be easily plugged into previous optimization-based methods and introduces realistic details on the coarse shapes to improve the visualized quality.

In addition, we also compare the reconstruction quality in the video sequences and further demonstrate that, given the coarse shape obtained by existing optimization-based methods, the proposed {\module} has the advantage of achieving realistic detailed results based on the existing coarse shape. As Fig.~\ref{fig:supp_ft_detail_video1}-\ref{fig:supp_ft_detail_video3} show, we perform reconstruction on the video sequences by comparing Dense~\citesupp{wood2022densesupp} with/without our {\module} to prior art~\citesupp{feng2021learningsupp,danvevcek2022emocasupp}. Fig.~\ref{fig:supp_ft_detail_video1}-\ref{fig:supp_ft_detail_video3} demonstrate that our {\module} significantly improve the visualized quality compared to the coarse shape from~\citesupp{wood2022densesupp}. Compared to prior art~\citesupp{feng2021learningsupp,danvevcek2022emocasupp}, our {\module} captures subtle and realistic details and outperforms previous methods by a large margin.

\subsection{Detail Animation Comparisons}

While previous state-of-the-art methods~\citesupp{feng2021learningsupp,danvevcek2022emocasupp} directly concatenate the person-specific identity features with expression-aware features and decode them into a displacement map. In our ablation studies, we have demonstrated that simply predicting the dynamic details is rather challenging to achieve satisfactory results (see Fig.~\ref{fig:abldynamic}). Here we present additional comparisons on detail animation to justify our claims.

\myparagraph{Qualitative Comparisons.} In Fig.~\ref{fig:supp_animation} and Fig.~\ref{fig:supp_animation2}, we manipulate the static and/or dynamic details of the source image by assigning the static and/or dynamic codes from the driving image. Specifically, the static coefficient in {\name} (or called ``detail code'' in DECA~\citesupp{feng2021learningsupp} and EMOCA~\citesupp{danvevcek2022emocasupp}) is encoded from the driving image. The dynamic factor is based on the expression coefficient in our {\name}, while in DECA~\citesupp{feng2021learningsupp} and EMOCA~\citesupp{danvevcek2022emocasupp}, it is based on the ``expression parameter'' and ``jaw pose''.
We present the comparisons of the animation results in Fig.~\ref{fig:supp_animation} and Fig.~\ref{fig:supp_animation2}.

We can see that the details are not well decoupled in previous methods.
For example, when we manipulate the static factor to the source image and the driving image is a young child ({\eg}, second column), the output shape should have presented ``young'' details.
However, results in previous methods still exhibit ``noisy'' details, which correspond to ``old'' wrinkles. It demonstrates that such implicit learning is hard to decouple the static and dynamic factors well.

As a comparison, our novelty and insights lie in the essence of 3DMMs that simplify the 2D-to-3D difficulty by statistical models.
We successfully reconstruct plausible static and dynamic details by simplifying such difficulty into feasible regression and interpolation tasks.
The details generated by {\name} are naturally decoupled into static and dynamic factors for animation. For example, if we animate the static factor, the facial details present variation among different age groups, and when we animate the dynamic factor, the expression-driven details are well transferred (see ``Ours'' in Fig.~\ref{fig:supp_animation} and Fig.~\ref{fig:supp_animation2} from left to right).

\myparagraph{User Study.} We also present another user study to investigate the objective evaluation from $5$ experienced artists in estimating the expression and detail transfer quality. 
More specifically, given the source images in Fig.~\ref{fig:supp_animation} and Fig.~\ref{fig:supp_animation2} (noted as subject I and subject II in Tab.~\ref{tab:supp_user2}), we ask the artists to mark scores ranging from 1 to 5 (5 indicates the best score) for each driving sample.
The scores are evaluated based on the animation quality {\wrt} the static details and the dynamic expressions from the driving images. The driving images are classified into: A. the young group (1-5 images) and B. the elder group (6-10 images). The quantitative results are summarized in Tab.~\ref{tab:supp_user2}.

According to Tab.~\ref{tab:supp_user2}, we demonstrate that our reconstruction and animation results are better aligned with human perception. For the static factor, we can see the results of previous methods~\citesupp{feng2021learningsupp,danvevcek2022emocasupp} present an imbalanced distribution, {\ie}, when the driving images are young people's, the score is, in general, worse than that of the elder group. As for the expression transfer, previous methods reach worse scores when the driving images contain extreme expressions (corresponding to a larger standard deviation in Tab.~\ref{tab:supp_user2}). As a comparison, our method presents higher scores with smaller variances among different age groups, which demonstrates the power of our model in transferring novel expressions and details.

\section{Limitations \& Future Work}
\label{sec.supp_limit}

This paper proposes a novel approach to reconstructing animatable details from monocular images. While we manage to synthesize realistic details and demonstrate higher accuracy compared to previous state-of-the-art, our work still has limitations. We pinpoint these challenges in the 3D face community and leave them for future work.

\myparagraph{Facial Appearance Model.}
We use the vanilla albedo 3DMM to linearly represent the facial appearance. While we focus more on the geometry shape, such albedo inherently lacks details and indirectly influences the training of {\name}. In the future, we plan to integrate the diffuse model and spectral model to present high-fidelity facial appearance and extend our {\name} with photo-realistic textures.

\myparagraph{Reconstruction Quality.}
While we achieve state-of-the-art reconstruction quality in the REALY~\cite{REALY} benchmark in terms of the overall quality in Tab.~\ref{tab:benchmark}. We notice {\name} does not perform the best in the mouth and cheek, which are highly emotional and structural regions. To address this problem, we leave it for future work to incorporate the emotion-aware perceptual loss and structure-aware constraints in these regions to further improve the reconstruction quality of {\name}.

\myparagraph{Displacement Prior.}
We demonstrate the necessity to leverage the statistical model to constrain the displacement distribution. However, due to the high expense of capturing large-scale and high-quality displacements for training a non-linear model. We choose the common practice of linear PCA model to build displacement bases for it is easy to implement and data amount friendly.
We trade off the learning difficulty and personalized details ({\eg}, nevus) through the statistical model. Therefore, it is still challenging to recover pore-level details.
In addition, we notice that the imbalanced data ({\ie}, a majority of young scans and images with fewer children and elders) also influence the representation of our model. In the future, we plan to capture and synthesize more class-balanced data to train a non-linear model and leverage the versatile generative models to reconstruct high-resolution displacement maps for more realistic 3D faces.

\myparagraph{Evaluation on Facial Animation.}
We present the visualized comparisons in transferring facial expressions and their details. However, we can only refer to quantitative comparisons for missing such a benchmark to estimate the transfer accuracy. We leave it for future work to construct a benchmark to evaluate the quality of expression transfer.

\section{Potential Social Impact} \label{sec.supp_social}

While this paper successfully reconstructs 3D shapes with animatable details from the monocular images, it is not intended to create content that is used to mislead or deceive. Therefore, this paper does not raise disinformation or immediate security concerns.
However, like other related 3D face reconstruction and animation techniques, it could still potentially be misused for impersonating humans. We condemn any behavior to create misleading or harmful content of real persons.
We encourage researchers in the 3D face community to consider the questions about preventing privacy disclosure before applying the model to the real world.

\balance{
{\small
\bibliographystylesupp{ieee_fullname}
\bibliographysupp{supp}
}
}

\end{document}